\documentclass{article} 
\usepackage[preprint]{colm2026_conference}

\usepackage{microtype}
\usepackage{hyperref}
\usepackage{url}
\usepackage{booktabs}
\usepackage{mathtools}
\usepackage{dblfloatfix}

\usepackage{subcaption}
\usepackage[most]{tcolorbox}
\usepackage{xcolor}
\usepackage{multirow}
\usepackage[bottom]{footmisc}

\usepackage{algorithm}
\usepackage{algpseudocode}
\usepackage{amssymb}

\usepackage{etoc}

\usepackage{lineno}
\usepackage{wrapfig}

\usepackage{tikz}     
\usepackage{mdframed} 
\usepackage{color-edits}
\usepackage{xcolor}
\addauthor{kb}{purple}

\definecolor{darkblue}{rgb}{0, 0, 0.5}
\hypersetup{colorlinks=true, citecolor=darkblue, linkcolor=darkblue, urlcolor=darkblue}
\definecolor{myblue}{RGB}{31,119,180}
\definecolor{mygray}{RGB}{110,110,110}

\gdef\Sepline{%
  \par\noindent\makebox[\linewidth][l]{%
  \hspace*{-\mdflength{innerleftmargin}}%
   \tikz\draw[thick,dashed,gray!60] (0,0) --%
        (\textwidth+\the\mdflength{innerleftmargin}+\the\mdflength{innerrightmargin},0);
  }\par\nobreak}

\newcommand{\name}{$p1$}

\title{\name: Better Prompt Optimization with Fewer Prompts}

\author{Zhaolin Gao$^1$\thanks{Correspondence to zg292@cornell.edu} , Yu (Sid) Wang$^2$, Bo Liu$^2$, Thorsten Joachims$^1$, Kianté Brantley$^3$, Wen Sun$^4$ \\
$^1$ Cornell University, $^2$ Microsoft, $^3$ Harvard University, $^4$ Databricks AI Research \\
}

\begin{document}

\ifcolmsubmission
\linenumbers
\fi

\maketitle

\begin{abstract}

Prompt optimization improves language models without updating their weights by searching for a better system prompt, but its effectiveness varies widely across tasks. We study what makes a task amenable to prompt optimization. We show that the reward variance across different system prompts can be decomposed into two components: variance \emph{among responses}, which captures generation stochasticity, and variance \emph{among system prompts}, which captures differences in system prompt quality. Prompt optimization succeeds when variance among system prompts is sufficiently large, but fails when variance among responses dominates the variance of the system prompts. Surprisingly, we further show that scaling to more user prompts can hurt optimization by reducing variance among system prompts, especially on heterogeneous datasets where different user prompts favor different system prompts. Motivated by this insight, we propose \name, a simple user prompt filtering method that selects a small subset of user prompts with high variance across candidate system prompts. This subset of user prompts allows one to distinguish a good system prompt from a bad one, making system optimization easier. Experiments on reasoning benchmarks show that \name{} substantially improves prompt optimization over training on the full dataset and outperforms strong baselines such as GEPA. Notably, training on only two prompts from AIME 24 yields a system prompt that generalizes well to other reasoning benchmarks.

\end{abstract}

\begin{figure}[b]
    \centering
    \includegraphics[trim={0 0 0 0}, clip, width=\textwidth]{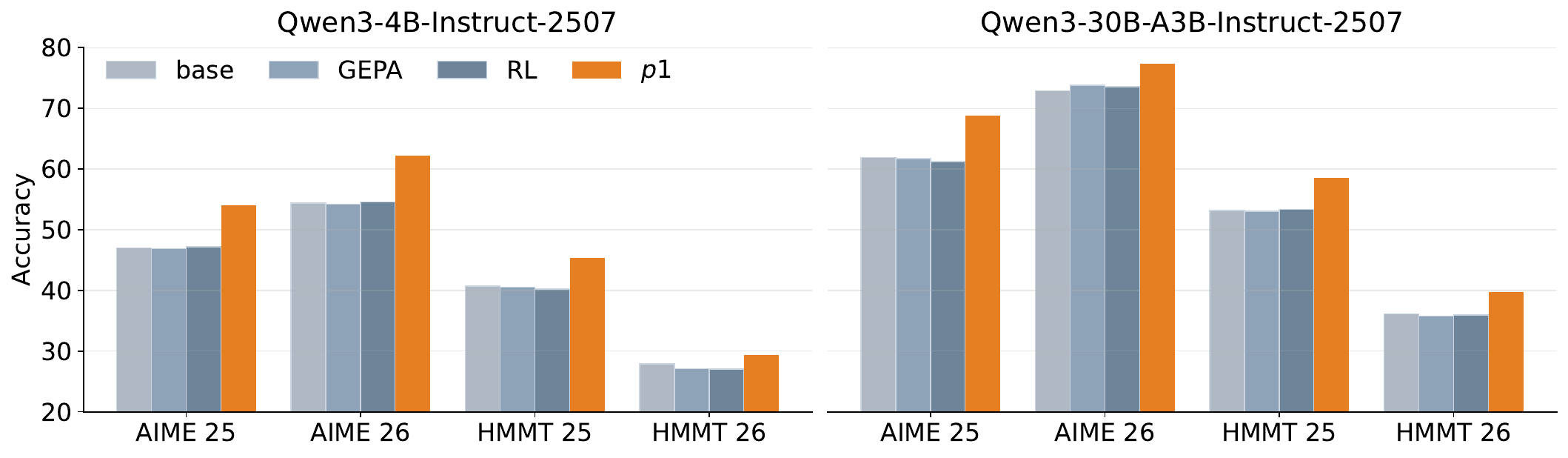}
    \vskip -0.2cm
    \caption{Comparison of \name{} against the base model and baseline methods. For all methods, the system prompt is optimized based on AIME 24 and Qwen3-4B-Instruct-2507, and directly applied to these benchmarks and to Qwen3-30B-A3B-Instruct-2507. The results are averaged over $64$ generations per user prompt.}
    \label{fig:fig1}
\end{figure}

\section{Introduction}

System prompts have become a central interface for steering large language models (LLMs). A well-designed system prompt can substantially improve task performance by shaping the model's reasoning style, formatting behavior, and adherence to instructions, all without modifying model weights~\citep{zhou2023largelanguagemodelshumanlevel,yang2024largelanguagemodelsoptimizers}. This makes prompt optimization an appealing alternative to full model training. Recent work has therefore explored automatic methods for optimizing prompts, including evolutionary search~\citep{fernando2023promptbreederselfreferentialselfimprovementprompt,guo2025evopromptconnectingllmsevolutionary,agrawal2026gepareflectivepromptevolution,liu2026evoxmetaevolutionautomateddiscovery}, and reinforcement learning (RL)~\citep{deng2022rlpromptoptimizingdiscretetext,kwon2024stablepromptautomaticprompttuning,xiao2025promptmiimetalearninginstructioninduction}. In this paper, we focus on the RL setting where a prompt-generation policy proposes candidate system prompts and the reward is the accuracy of a frozen LLM under each candidate system prompt. 

Despite the simplicity of prompt optimization, the performance is strikingly inconsistent. On some tasks, optimized system prompts yield clear gains; on others, optimization barely improves even with substantial compute. We investigate the underlying mechanisms that govern this inconsistency. We demonstrate that the reward variance across different system prompts can be decomposed into two distinct components: \textbf{variance among responses} (e.g, math solutions), which captures the inherent generation stochasticity under a fixed system prompt, and \textbf{variance among system prompts}, which captures the true expected reward differences between system prompts. We show that prompt optimization is effective only when the latter component is sufficiently large. For example, on instruction-following benchmarks (e.g., IFBench~\citep{pyatkin2025generalizingverifiableinstructionfollowing}), rewards are highly sensitive to the system prompt, creating a clean optimization signal. Conversely, on complex reasoning benchmarks (e.g., AIME~\citep{balunovic_srimatharena_2025}), the optimization signal is heavily obscured because the variance among responses dominates the true differences between system prompts.

Moreover, our analyses reveal a counterintuitive phenomenon that \textbf{increasing the size of the dataset reduces the variance among system prompts}. Since different user prompts (e.g, math questions) can favor different system prompts, averaging over a larger dataset causes these preferences to cancel out. Candidate system prompts begin to appear statistically identical in terms of their expected reward, diluting the signal for automatic prompt optimization. This effect is especially severe on heterogeneous tasks such as mathematical reasoning, where system prompts that help one example may harm another. By contrast, on more homogeneous tasks such as instruction following, a good system prompt tends to help many examples consistently, making prompt optimization feasible over a large dataset. 

Motivated by this insight, we propose \textbf{\name{}, a simple and effective user-prompt filtering method that improves prompt optimization while using fewer training examples}. Instead of optimizing over the full training set, \name{} selects a small subset of user prompts that exhibit high reward variance across candidate system prompts. Training on this filtered subset strengthens the optimization signal by focusing on prompts that best distinguish strong system prompts from weak ones. We evaluate \name{} on both instruction-following and reasoning benchmarks. Our results show that standard prompt optimization can fail on heterogeneous reasoning datasets even with substantial compute, whereas filtering yields substantial improvements. Notably, \textbf{training on just two prompts from AIME 24 produces a system prompt that generalizes well to other reasoning benchmarks}. The learned prompts also transfer beyond the training setting. As shown in Fig.~\ref{fig:fig1}, system prompts optimized for Qwen3-4B-Instruct-2507 also improve the larger Qwen3-30B-A3B-Instruct-2507 and boost performance on additional reasoning benchmarks not seen during training.


\section{Problem Setup}

Let $x'$ denote a system prompt and $x$ denote a user prompt (e.g., a math question). Given the tuple $(x', x)$, we autoregressively sample a response $y$ from a language model policy $\pi$, $y \sim \pi(\cdot \mid x', x)$. We define a binary reward function $r(x, y) \in \{0, 1\}$, where $r(x, y) = 1$ indicates a correct response and $0$ otherwise. The objective of prompt optimization is to find a system prompt $x'$ that maximizes the expected reward across a given dataset $\mathcal{D}$:
\begin{align}
    \max_{x'} \mathbb{E}_{x \sim \mathcal{D},\, y \sim \pi(\cdot \mid x', x)} \left[ r(x, y) \right].
    \label{eq:main_obj_1}
\end{align}

Following prior work, we cast prompt optimization as a reinforcement learning (RL) problem~\citep{deng2022rlpromptoptimizingdiscretetext,kwon2024stablepromptautomaticprompttuning,xiao2025promptmiimetalearninginstructioninduction}. Let $s$ denote a meta-prompt, and let $\pi'$ denote a system prompt generation policy. Our objective is to improve $\pi'$ such that it produces increasingly effective system prompts:
\begin{align}
    \max_{\pi'} \mathbb{E}_{x' \sim \pi'(\cdot \mid s),\, x \sim \mathcal{D},\, y \sim \pi(\cdot \mid x', x)} \left[ r(x, y) \right].
    \label{eq:main_obj_2}
\end{align}

Equivalently, we can define $r(x') = \mathbb{E}_{x \sim \mathcal{D},\, y \sim \pi(\cdot \mid x', x)} [r(x, y)]$ as the expected reward of a system prompt $x'$ where $\pi$ is fixed throughout training, simplifying our objective to maximizing $\mathbb{E}_{x' \sim \pi'(\cdot \mid s)} [r(x')]$. While any RL algorithm could be used, we adopt a purely on-policy variant of GRPO~\citep{shao2024deepseekmath,deepseekai2025deepseekr1}, without KL regularization to a reference policy $\pi_{\mathrm{ref}}$~\citep{yu2025dapoopensourcellmreinforcement} and without standard-deviation-based advantage normalization~\citep{liu2025understandingr1zeroliketrainingcritical}. At step $t$, we maximize the following objective:
\begin{align}
    \mathbb{E}_{x' \sim \pi'_t(\cdot \mid s)} \left[
    \frac{1}{|x'|} \sum_{l=1}^{|x'|}
    \frac{\pi'(x'_l \mid s, x'_{<l})}{\pi'_t(x'_l \mid s, x'_{<l})}
    \bigl(r(x') - V^{\pi'_t}(s)\bigr)
    \right],
    \label{eq:main_obj}
\end{align}
where $x'_l$ denotes the $l$-th token of the generated system prompt $x'$, and $V^{\pi'_t}(s) := \mathbb{E}_{x' \sim \pi'_t(\cdot \mid s)} [r(x')]$ serves as the value baseline. We select this formulation because it provides a clean RL objective that is equivalent to RLOO~\citep{ahmadian2024basicsrevisitingreinforcestyle} where the gradient matches with standard policy gradient~\citep{gao2025promptcurriculumlearningefficient}.


\section{Prompt Learnability}

In this section, we present a set of preliminary experiments to study the learnability of user prompts, i.e., which classes of user prompts can be optimized effectively with better system prompts and which appear substantially more difficult to improve. To optimize the objective in Eq.~\ref{eq:main_obj}, for each step, we sample $N$ system prompts $x'$ from $\pi'_t(\cdot \mid s)$. For each sampled system prompt and each user prompt $x$ in the dataset $\mathcal{D}$, we then draw $M$ responses $y$, yielding a total of $KM$ sampled responses per system prompt, where $K$ is the size of the dataset $\mathcal{D}$, and a total of $KNM$ sampled responses per step. The reward assigned to $x'$ is estimated using the Monte Carlo estimator $\hat{r}(x') = \frac{1}{KM}\sum_{k=1}^{K}\sum_{m=1}^{M} r(x_k, y_k^m)$. This estimator approximates the correctness achieved by system prompt $x'$ over the dataset, marginalizing over the stochasticity of the response policy $\pi$ through repeated sampling.

\subsection{Initial Investigations}
\label{sec:initial_investigation}

\textbf{Experimental Setup.} We conduct experiments on IFBench~\citep{pyatkin2025generalizingverifiableinstructionfollowing} and AIME~\citep{balunovic_srimatharena_2025}. IFBench is a benchmark designed to evaluate language models’ ability to follow precise human instructions, particularly those involving strict output constraints. Its test set contains 58 out-of-distribution constraints. For training, we use a subset of $64$ questions from IF-RLVR, which is the training set of IFBench, and evaluate on the IFBench, which ensures that system prompt optimization is not performed directly on the unseen constraints used for evaluation. The reward function is binary, where it assigns 1 if the model satisfies all specified constraints and 0 otherwise. For AIME, a competition-level math dataset, we use all $30$ questions from AIME 2024 for training and use AIME 2025 for evaluation. We adopt a rule-based reward function based on \texttt{math-verify}~\citep{mathverify2024}, which assigns a reward of 1 to correct answers and 0 to incorrect answers or generations that exceed the context limit.

We use Qwen3-4B-Instruct-2507~\citep{qwen3technicalreport} as both $\pi$ and $\pi'$, where $\pi$ is frozen as the response policy, and $\pi'$ is updated during training. We set the generation length of $\pi'$ to 4{,}096 tokens, and the generation length of $\pi$ to 2{,}048 tokens for IFBench and 16{,}384 tokens for AIME. All experiments are implemented using \textsc{Verl}~\citep{Sheng_2025}. For evaluation, we compute the mean accuracy across $4$ system prompts ($N=4$) generated from the trained $\pi'$ by generating $1$ or $8$ responses per user prompt ($M$) for each system prompt on IFBench or AIME, respectively. The experiments are conducted on 4 H100 GPUs, where we use 1 GPU for updating and generating system prompts from $\pi'$ and 3 GPUs for generating responses from $\pi$. Additional training details, including the prompt format and the content for meta prompt $s$, are provided in Appendix~\ref{app:exp_detail} with generated examples in Appendix~\ref{app:system_prompts}.

\begin{figure}[t]
    \centering
    \includegraphics[trim={0 0 0 0}, clip, width=\textwidth]{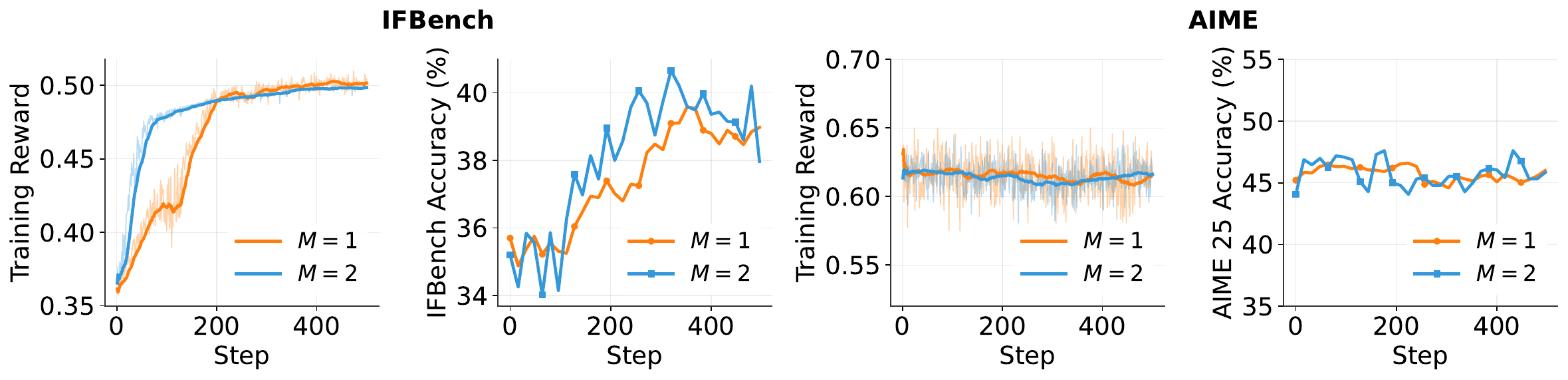}
    \vskip -0.3cm
    \caption{Training reward and evaluation accuracy on IFBench and AIME with $M\in\{1, 2\}$.}
    \label{fig:ifbench_aime}
    \vskip -0.3cm
\end{figure}

\textbf{Contrasting behavior on IFBench and AIME.} We set $N=16$ and ablate $M \in \{1,2\}$. We use small values of $M$ since gathering the reward of each system prompt is expensive, requiring $KM$ responses per system prompt. Even with $M=2$, reward collection takes around 1{,}400 seconds per training step.
The training reward and evaluation accuracy are shown in Fig.~\ref{fig:ifbench_aime}. On IFBench, the training reward increases steadily until convergence, and the evaluation accuracy improves accordingly. In contrast, on AIME, the training reward remains flat, and the evaluation accuracy exhibits a similar pattern. We also evaluate the evolution-based prompt optimization method GEPA~\citep{agrawal2026gepareflectivepromptevolution}, which also improves performance on IFBench but not on AIME. This difference raises a natural question: \textbf{\textit{what properties make a user prompt more amenable to prompt optimization?}}

\subsection{Variances for One Prompt}
\label{sec:variance_for_one_prompt}

We consider a simplified setting with $K=1$, i.e., the dataset $\mathcal{D}$ contains only a single user prompt. Since the reward is binary, the reward induced by a fixed system prompt $x'$ is a Bernoulli distribution with mean $p :=r(x') = \mathbb{E}_{y \sim \pi(\cdot \mid x', x)}[r(x,y)]$. To estimate the reward of $x'$, we sample $M$ responses $y^1,\dots,y^M$ from $\pi(\cdot \mid x', x)$ and compute the empirical mean $\hat{r}(x') = \frac{1}{M}\sum_{m=1}^M r(x, y^m)$, where $r(x, y^m) \overset{\text{i.i.d.}}{\sim} \mathrm{Bernoulli}(p)$. Since it is a Bernoulli distribution, we can easily derive the variance of $\hat{r}(x')$: $\mathrm{Var}(\hat{r}(x')) = \frac{p(1-p)}{M}$.

For $N$ generated system prompts $x'_1, \dots, x'_N$, let the reward associated with $x'_n$ have mean $p_n$, such that {\color{myblue} $\mathrm{Var}(\hat{r}(x'_n)) = \frac{p_n(1-p_n)}{M}$}. Denote the variance across the rewards of these system prompts as $\mathrm{Var}(\hat{r}) = \frac{1}{N}\sum_{n=1}^N \bigl(\hat{r}(x'_n)-\bar r \bigr)^2$ where $\bar r = \frac{1}{N}\sum_{n=1}^N \hat{r}(x'_n)$. Then the expected variance over the randomness of the sampled responses can be formulated as
\begin{align}
    \mathbb{E}[\mathrm{Var}(\hat{r})]
    &=
    \underbrace{{\color{myblue}\frac{N-1}{N^2}\sum_{n=1}^N \frac{p_n(1-p_n)}{M}}}_{\text{\textbf{{\color{myblue}among responses}}}}
    +
    \underbrace{{\color{mygray}\frac{1}{N}\sum_{n=1}^N (p_n-\bar p)^2}}_{\text{\textbf{{\color{mygray}among system prompts}}}},
    \label{eq:one_prompt_var}
\end{align}
where $\bar p = \frac{1}{N}\sum_{n=1}^N p_n.$ The proof is included in Appendix~\ref{app:exp_var_proof}. This decomposition shows that the expected variance across rewards for system prompts consists of two terms: the first term arises from Monte Carlo estimations \textbf{among responses}, and the second term reflects the true differences in expected reward \textbf{among system prompts}. Ideally, we want to decrease the first term as much as possible through repeated sampling (i.e., increasing $M$) to have a clean reward signal. 

\textbf{Experiment Setup.} To analyze the variance characteristics of IFBench and AIME, we sample $N=16$ system prompts from Qwen3-4B-Instruct-2507 ($\pi'$ at step $0$). For each system prompt and a user prompt, we generate $M=128$ responses to estimate $\hat{r}(x'_n)$ and $\mathrm{Var}(\hat{r})$ for IFBench training set and AIME 24. We estimate the variance among responses using the mean over the 128 sampled responses as a proxy for $p_n$. The variance among system prompts is computed by subtracting the response variance from the total variance $\mathrm{Var}(\hat{r})$.

\begin{figure}[t]
    \centering
    \includegraphics[trim={70 300 70 0}, clip, width=\textwidth]{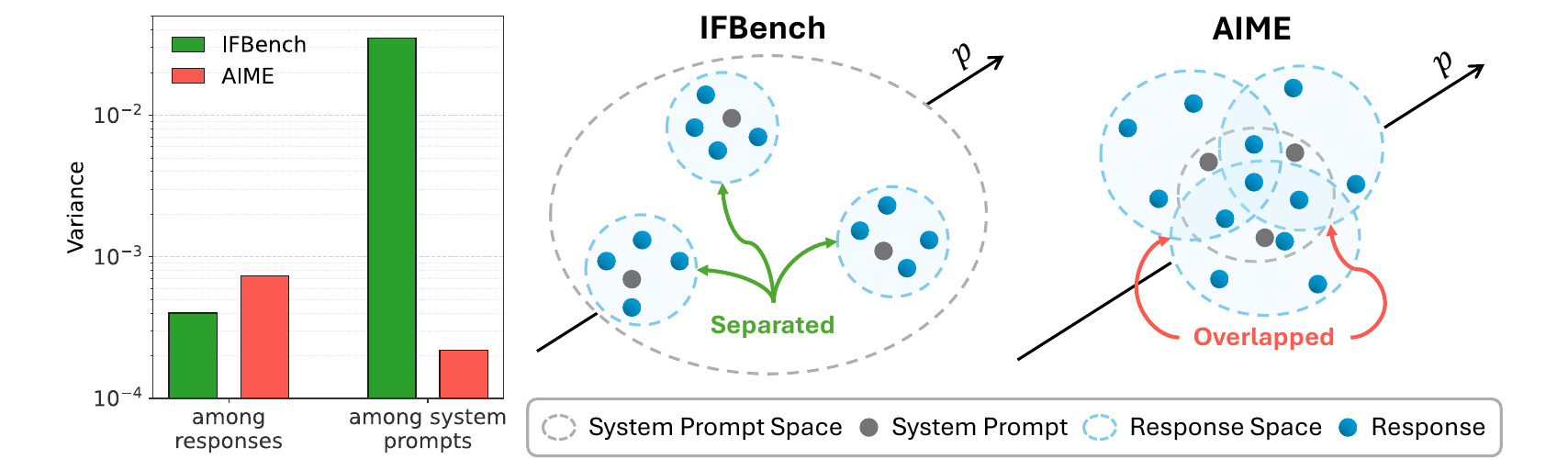}
    \vskip -0.2cm
    \caption{Variances across responses and system prompts for IFBench training set and AIME 24. On IFBench, the variance in reward across system prompts is substantially larger than the variance across sampled responses. In contrast, on AIME, the variance across system prompts is smaller than the variance across responses. The visualizations provide an intuitive illustration of this distinction. On IFBench, different system prompts are more clearly separated, making it easier to identify which system prompt is better, whereas on AIME, the small variance makes such distinctions much harder to detect.}
    \label{fig:ifbench_aime_variance}
    \vskip -0.3cm
\end{figure}

\textbf{Sensitivity to System Prompts.} As shown in Fig.~\ref{fig:ifbench_aime_variance}, IFBench exhibits substantially higher variance across system prompts than AIME. In other words, IFBench is highly sensitive to system prompts where different system prompts induce responses with significantly different rewards. In contrast, AIME is remarkably less sensitive to the system prompt, with variance being dominated by the stochasticity of the generation process itself. This creates a challenging environment for RL. On IFBench, the clean separation of reward spaces allows the policy to easily identify and optimize toward better prompts. On AIME, the high generation noise and low prompt sensitivity cause the response spaces of different system prompts to overlap heavily, reducing the learning signal.

\begin{wrapfigure}{r}{0.4\textwidth}
    \centering
    \vspace{-0.3cm} 
    \includegraphics[trim={0 0 0 0}, clip, width=0.4\textwidth]{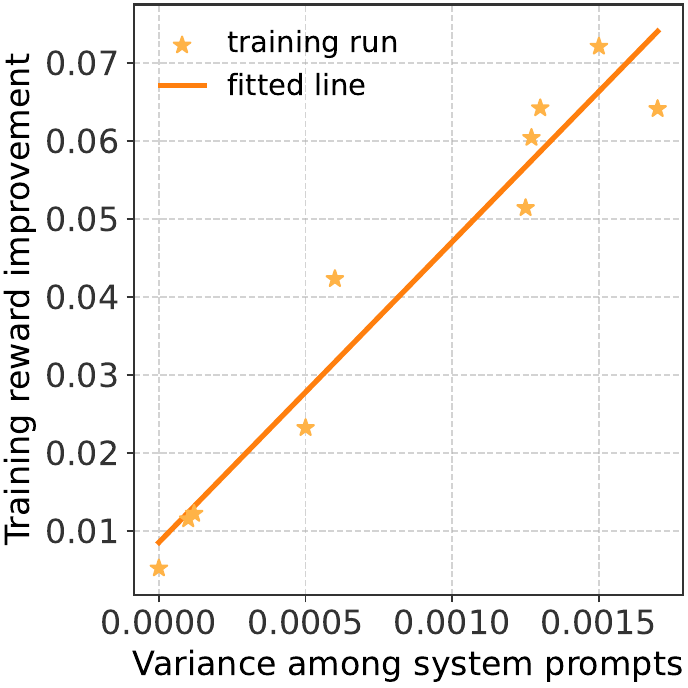}
    \vspace{-0.5cm}
    \caption{Variance among system prompts vs. training reward improvement when training on one AIME prompt.}
    \label{fig:variance_improvement}
    \vspace{-0.3cm}
\end{wrapfigure}

\textbf{Prompt learnability correlates with variance among system prompts.} To investigate the effect of variance in prompt optimization, we perform prompt optimization on a single AIME 24 prompt at a time, across 10 different prompts. We follow the same setup as in Sec.~\ref{sec:initial_investigation}, except that we set $M=32$ since training is performed on only one user prompt (i.e., $K=1$). The results are shown in Fig.~\ref{fig:variance_improvement}, where each point represents the average over three independent runs. We measure improvement as the difference between the final and initial training reward. These 10 AIME prompts exhibit different levels of variance among system prompts, and the resulting improvement in training reward shows a clear linear correlation with this variance. In other words, prompts that are more sensitive to the choice of system prompt tend to benefit more from prompt optimization, yielding larger gains in reward. 

However, the preceding analyses present an apparent discrepancy that, while training on the full 30-question AIME 24 dataset yielded no improvement in training reward, optimizing on a single AIME question with sufficient variance could successfully increase the reward. This discrepancy is counterintuitive, given that the 30-question set includes those high-variance prompts that proved to be learnable by themselves. To understand why increasing size of the dataset appears to hinder rather than help the optimization process, we now extend our variance analysis from the single-prompt setting ($K=1$) to the multiple prompts ($K > 1$).

\subsection{Variances for Multiple Prompts}
\label{sec:variance_for_multiple_prompt}

Given a dataset $\mathcal{D}=\{x_k\}_{k=1}^K$ of size $K$, and $N$ generated system prompts $x'_1,\dots,x'_N$, define $p_n^k := \mathbb{E}_{y \sim \pi(\cdot \mid x'_n, x_k)}[r(x_k,y)]$ as the success probability of system prompt $x'_n$ on user prompt $x_k$. The expected reward of $x'_n$ over the dataset is then $r(x'_n)=\frac{1}{K}\sum_{k=1}^K p_n^k$. We estimate this quantity using $\hat r(x'_n)=\frac{1}{KM}\sum_{k=1}^K\sum_{m=1}^M r(x_k,y_k^m)$, where $r(x_k,y_k^m)\overset{\text{i.i.d.}}{\sim}\mathrm{Bernoulli}(p_n^k)$, and can easily derive the variance as {\color{myblue}$\mathrm{Var}(\hat{r}(x'_n)) = \frac{1}{K} \sum_{k=1}^K \frac{p^k_n(1-p^k_n)}{KM}$}. Similarly, we can denote the variance across system prompts rewards as $\mathrm{Var}(\hat{r})$, and the expected variance over the randomness of the sampled responses can be formulated as:
\begin{align}
    \mathbb{E}[\mathrm{Var}(\hat{r})]
    &=
    \underbrace{
    {\color{myblue}
    \frac{N-1}{N^2}
    \sum_{n=1}^N
    \left(
    \frac{1}{K} \sum_{k=1}^K \frac{p^k_n(1-p^k_n)}{KM}
    \right)}
    }_{{\color{myblue}\text{\textbf{among responses}}}}
    +
    \underbrace{
    {\color{mygray}
    \frac{1}{N}\sum_{n=1}^N (p_n-\bar p)^2}
    }_{{\color{mygray}\text{\textbf{among system prompts}}}},
    \label{eq:multi_prompt_var}
\end{align}
where $p_n=r(x'_n)$ and $\bar p=\frac{1}{N}\sum_{n=1}^N r(x'_n)$. The proof is provided in Appendix~\ref{app:exp_var_proof_multi}. Similarly, the above expected variance can also be decomposed into two terms, which are among responses and among system prompts. Different from Eq.~\ref{eq:one_prompt_var}, we now have $K$ in the first term, and the variance among responses is inversely correlated with both $K$ and $M$.

\textbf{Experiment Setup.} To analyze the variance, we follow the same setup in Sec.~\ref{sec:variance_for_one_prompt} and vary $K$ and $M$. For each $K$, we consider all combinations of $K$ prompts drawn from the 30 AIME prompts or 64 IFBench training prompts. For each $M$, we uniformly sample $M$ responses from the 128 responses and repeat this process for 100 trials. As in Sec.~\ref{sec:variance_for_one_prompt}, we use the mean over 128 sampled responses as a proxy for $p_n$ for estimating the variance among responses.

\begin{figure}[t]
    \centering
    \includegraphics[trim={0 0 0 0}, clip, width=\textwidth]{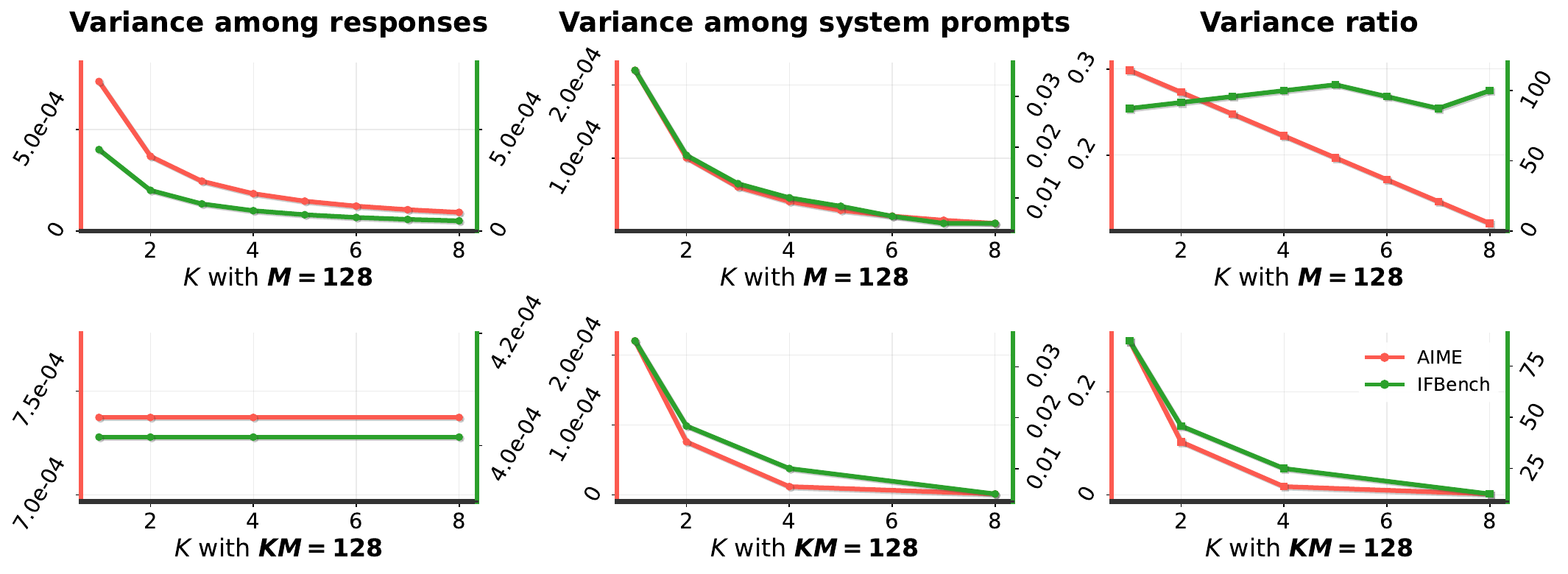}
    \vskip -0.2cm
    \caption{Variance among responses and among system prompts for AIME and IFBench. The first row shows results with varying $K$ while fixing $M=128$. The second row shows results with varying both $K$ and $M$ while keeping $KM=128$ fixed. The last column reports the ratio of the variance among system prompts to the variance among responses. Note that each line is plotted with its own y-axis scale.}
    \label{fig:variance_grid}
    \vskip -0.3cm
\end{figure}

\textbf{Variance among responses scales as $1/(KM)$.} Figure~\ref{fig:variance_grid} reports the results. In the first row, we vary $K$ while fixing $M=128$; in the second row, we vary both $K$ and $M$ while keeping $KM=128$ constant. The first column shows that when $M$ is fixed, increasing $K$ monotonically reduces the variance among responses. In contrast, when $KM$ is held constant, the response variance remains unchanged across different choices of $K$ and $M$. This behavior is exactly consistent with Eq.~\ref{eq:multi_prompt_var} where the noise introduced by response stochasticity is controlled primarily by the total sample budget $KM$.

\textbf{Increasing the size of the dataset reduces the variance among system prompts.} The second column of Fig.~\ref{fig:variance_grid} shows that, as $K$ increases, the variance among system prompts decreases for both AIME and IFBench, regardless of whether $M$ is fixed or adjusted. This indicates that different user prompts often favor different system prompts. A system prompt that improves performance on one example may hurt performance on another, and averaging across a larger and more diverse dataset makes the true reward $p_n$ of the system prompt $x'_n$ increasingly similar to the rewards of other system prompts. As a result, the distinction between good and bad system prompts becomes less clear as the dataset gets larger. 

\textbf{Prompt optimization works better on homogeneous datasets.} The third column in Fig.~\ref{fig:variance_grid} reports the ratio between the variance among system prompts and the variance among responses. This ratio can be viewed as a signal-to-noise ratio for prompt optimization. Larger values indicate that differences between system prompts are easier to detect relative to sampling noise. On AIME, this ratio decreases as $K$ grows, even when $M$ is kept fixed. In other words, simply adding more user prompts does not strengthen the optimization signal. Maintaining the same signal-to-noise ratio would require increasing the sampling budget faster than linearly with dataset size, which is computationally expensive in practice. 
The situation is even worse when $KM$ is fixed, since increasing $K$ necessarily reduces $M$, causing the signal-to-noise ratio to deteriorate even more rapidly (the lower plot of the third column). By contrast, IFBench behaves much more favorably. When $M$ is fixed, the signal-to-noise ratio remains constant as $K$ increases, indicating that prompts in IFBench are more \emph{homogeneous}: system prompts that work well on one user prompt tend to also work well on other user prompts. In such settings, scaling the dataset provides a consistent learning signal, and prompt optimization remains effective as the dataset increases.

\section{\texorpdfstring{\name: Prompt Filtering for Better Prompt Optimization}{NAME: Prompt Filtering for Better Prompt Optimization}}
\label{sec:prompt_filtering}

The analysis in Sec.~\ref{sec:variance_for_one_prompt} and~\ref{sec:variance_for_multiple_prompt} suggests that prompt optimization is effective only when the reward signal remains sufficiently distinguishable across candidate system prompts, and, as dataset size increases, the reward signal decreases. We propose \name, a simple data selection strategy that retains only a small subset of user prompts whose rewards exhibit large variance across system prompts.

Following Eq.~\ref{eq:multi_prompt_var}, for a subset $\mathcal{S} \subseteq \mathcal{D}$ with $|\mathcal{S}| = K_{\mathrm{top}}$, where $K_{\mathrm{top}}$ is a hyperparameter set to 2 by default, we estimate the variance among system prompts by computing $\hat r(x'_n)$ and $\mathrm{Var}(\hat r)$ using a sufficiently large value of $M$. We then estimate the variance among responses by using the empirical mean over $M$ sampled responses as a proxy for $p_n$. The variance among system prompts is obtained by subtracting the estimated response variance from $\mathrm{Var}(\hat r)$. We evaluate this quantity for all possible subsets and select the subset with the largest score.

Note that we do not directly use the approximated $p_n$ to compute the variance among system prompt, since it would be the exact same way as computing $\mathrm{Var}(\hat r)$ which also contains variance among responses. It may have a bias toward prompt with $p^k_n=0.5$ as they have the highest variance among responses. By explicitly estimating and subtracting the variance among responses, we obtain a cleaner estimate of the true signal relevant for prompt optimization. We also use the estimated variance among system prompts rather than a signal-to-noise ratio since the ratio is unstable and would bias selection toward easy or hard prompts where \(p_n^k\) is close to \(1\) or \(0\) as they have small variance among responses. Using the actual value of the variance yields a simpler and more stable selection criterion.

After filtering, we run the same prompt optimization procedure as in Sec.~\ref{sec:initial_investigation}, except that training is performed only on the selected subset. \name{} increases the average variance among system prompts within the retained data and yields a more homogeneous training set, thereby strengthening the reward signal. A pseudo-code is provided in Appendix~\ref{app:pseudocode}.

\section{Experiments}

\textbf{Models \& Datasets.} We use the same datasets as in Sec.~\ref{sec:initial_investigation}. For the system-prompt generator $\pi'$, we use Qwen3-4B-Instruct-2507 with a maximum generation length of 4{,}096 tokens. For the response model $\pi$, we consider either Qwen3-4B-Instruct-2507 (generation lengths of 2{,}048 tokens on IFBench and 16{,}384 tokens on AIME), or Qwen3-1.7B (generation lengths of 4{,}096 tokens on IFBench and 32{,}768 tokens on AIME). All experiments are conducted under a three-day time budget on four H100 GPUs. For system prompts trained on AIME, we additionally evaluate on AIME 2026, HMMT Nov 2025 \& Feb 2026~\citep{balunovic_srimatharena_2025}.

\textbf{Baselines.} We compare \name{} against GEPA~\citep{agrawal2026gepareflectivepromptevolution}, an evolution-based prompt optimization method, and RL-based prompt optimization (RL) applied to the full training set (RL is \name{} but with $K_{\mathrm{top}}=K$). For both RL and \name{}, we keep $K M$ roughly constant (when using a smaller $K_{\mathrm{top}}$ for \name{}, we increase $M$ accordingly). For GEPA, we either randomly split the training set into equally sized training and validation subsets, or use the subset filtered by \name{} for one of the training or validation subsets. Full details are provided in Appendix~\ref{app:exp}.

\textbf{\name{} outperforms GEPA and RL on reasoning benchmarks.} Table~\ref{tab:aime_hmmt_results} reports results on AIME and HMMT. We evaluate \name{} on two top-ranked subsets selected by our filtering procedure, with $K_{\mathrm{top}} \in \{1,2,4\}$, since there is inherent noise in estimating the variance. While both full-dataset RL and GEPA remain close to the base model (even when GEPA uses the top subset [1, 23] as one of its training or validation sets), \name{} achieves clear gains, indicating that standard prompt optimization struggles to obtain a useful learning signal on these heterogeneous reasoning tasks. When $K_{\mathrm{top}}=1$, \name{} improves the training reward but fails to generalize, overfitting to a single prompt. Using slightly larger subsets leads to much stronger generalization. In particular, training on prompts $[1,23]$ gives the best performance across all reasoning benchmarks, and $[17,27]$ also outperforms full-dataset RL. One possible explanation is that these subsets yield a cleaner signal, allowing $\pi'$ to move further from its initialization, whereas training on the full dataset is dominated by noise and remains close to the starting point. Although training on a subset may have limited generalization, it provides meaningful progress where training on the full dataset does not. We further evaluate the best system prompt learned from subset $[1,23]$ on Qwen3-30B-A3B-Instruct-2507, with results shown in Fig.~\ref{fig:fig1}. The gains transfer not only across reasoning benchmarks but also across models within the same family.

\begin{table*}[t]
\centering
\small
\resizebox{0.95\textwidth}{!}{%
\begin{tabular}{clccccccc}
\toprule
& \multirow{2}{*}{Method} & \multirow{2}{*}{$S$} & \multirow{2}{*}{$M$} & Training & \multirow{2}{*}{AIME 25} & \multirow{2}{*}{AIME 26} & \multirow{2}{*}{HMMT 25} & \multirow{2}{*}{HMMT 26} \\
&  &  &  & Reward Imp. &  &  &  &  \\
\midrule
\multirow{10}{*}{\rotatebox{90}{Qwen3-4B-Instruct-2507}}
& base  & /      & /  & /    & 47.03 & 54.38 & 40.68 & 27.89 \\
& GEPA  & 15/15      & /  & /    & 46.87 & 54.22 & 40.57 & 27.08 \\
& GEPA  & [1,23]/28  & /  & /  & 46.72 &  53.49 & 40.10 & 27.46 \\
& GEPA  & 28/[1,23]  & /  & /  & 46.35 &  53.91 & 40.21 & 27.32 \\
& RL    & all 30 & 1  & 0    & 47.24 & 54.58 & 40.26 & 27.04 \\
& \name & [4, 5, 17, 20] & 8  & 0.04 & 48.23 & 55.68 & 42.08 & 28.13 \\
& \name & [1, 23]        & 16 & 0.21 & \textbf{54.01} & \textbf{62.24} & \textbf{45.42} & \textbf{29.40} \\
& \name & [17, 27]       & 16 & 0.09 & \underline{50.10} & \underline{57.24} & \underline{42.34} & \underline{28.27} \\
& \name & [23]           & 32 & 0.07 & 47.81 & 55.21 & 41.25 & 27.98 \\
& \name & [25]           & 32 & 0.07 & 44.95 & 51.25 & 38.59 & 26.09 \\
\midrule
\multirow{10}{*}{\rotatebox{90}{Qwen3-1.7B}}
& base  & /               & /  & /    & 35.63 & 35.57 & 25.36 & \textbf{23.96} \\
& GEPA  & 15/15           & /  & /    & 34.17 & 35.10 & 25.16 & \underline{23.77} \\
& GEPA  & [5,26]/28  & /  & /   & 33.85 & 35.26 & 25.00 & 23.72 \\
& GEPA  & 28/[5,26]  & /  & /   & 34.22 & 35.52 & 25.52 & 23.58 \\
& RL    & all 30          & 1  & 0    & 35.10 & 35.89 & 24.90 & 23.44 \\
& \name & [5, 14, 20, 26] & 8  & 0.05 & 36.04 & 35.99 & 24.79 & 23.34 \\
& \name & [10, 11]        & 16 & 0.14 & \underline{36.10} & \underline{36.15} & \underline{26.93} & 23.58 \\
& \name & [5, 26]         & 16 & 0.13 & \textbf{36.98} & \textbf{37.60} & \textbf{27.08} & 23.48 \\
& \name & [5]             & 32 & 0.25 & 34.58 & 35.63 & 25.05 & 23.20 \\
& \name & [26]            & 32 & 0.29 & 34.11 & 33.23 & 24.38 & 22.96 \\
\bottomrule
\end{tabular}
}
\vspace{-0.2cm}
\caption{Performance on AIME and HMMT under different methods and selected subsets $S$. The subset $S$ is a set of indices (within []) or the number of prompts used for training. For GEPA, $S$ shows the (training prompts / validation prompts) split. The best performing method is highlighted in \textbf{bold} and the second best is \underline{underlined}.}
\vspace{-0.4cm}
\label{tab:aime_hmmt_results}
\end{table*}

\begin{wraptable}{r}{0.48\textwidth}
\vspace{-0.4cm}
\centering
\small
\resizebox{0.48\textwidth}{!}{%
\begin{tabular}{lcccc}
\toprule
\multirow{2}{*}{Method} & \multirow{2}{*}{$S$} & \multirow{2}{*}{$M$} & Training & \multirow{2}{*}{IFBench} \\
& & & Reward Imp. & \\
\midrule
\midrule
\multicolumn{5}{c}{Qwen3-4B-Instruct-2507} \\
\midrule
\midrule
base  & /  & /   & /    & 35.03 \\
GEPA  & 32/32  & /   & /    & 39.12 \\
RL    & 64 & 2   & 0.15 & 39.46 \\
\name  & 16 & 8   & 0.18 & 37.41 \\
\name  & 4  & 32  & 0.25 & 35.71 \\
\name  & 1  & 128 & 0.38 & 35.37 \\
\midrule
\midrule
\multicolumn{5}{c}{Qwen3-1.7B} \\
\midrule
\midrule
base  & /  & /   & /    & 24.49 \\
GEPA  & 32/32  & /   & /    & 30.95 \\
RL    & 64 & 2   & 0.05 & 31.97 \\
\name  & 16 & 8   & 0.09 & 29.25 \\
\name  & 4  & 32  & 0.13 & 26.53 \\
\name  & 1  & 128 & 0.27 & 24.83 \\
\bottomrule
\end{tabular}%
}
\vspace{-0.2cm}
\caption{Performance on IFBench under different methods, subset $S$ and $M$.}
\label{tab:ifbench_results}
\vspace{-0.4cm}
\end{wraptable}

\textbf{On IFBench, GEPA and RL perform similarly while \name{} is less effective.} Table~\ref{tab:ifbench_results} shows that both GEPA and RL achieve strong results on IFBench, whereas \name{} underperforms in this setting. As $K_{\mathrm{top}}$ decreases, training is performed on smaller subsets of user prompts, which causes overfitting. Although the training reward gain becomes larger for smaller $K_{\mathrm{top}}$, these gains do not translate into better evaluation performance on IFBench. This pattern suggests that, since IFBench is relatively homogeneous, it is more beneficial to use the full set, which has a strong learning signal while also yielding better generalization.

\begin{figure*}[t]
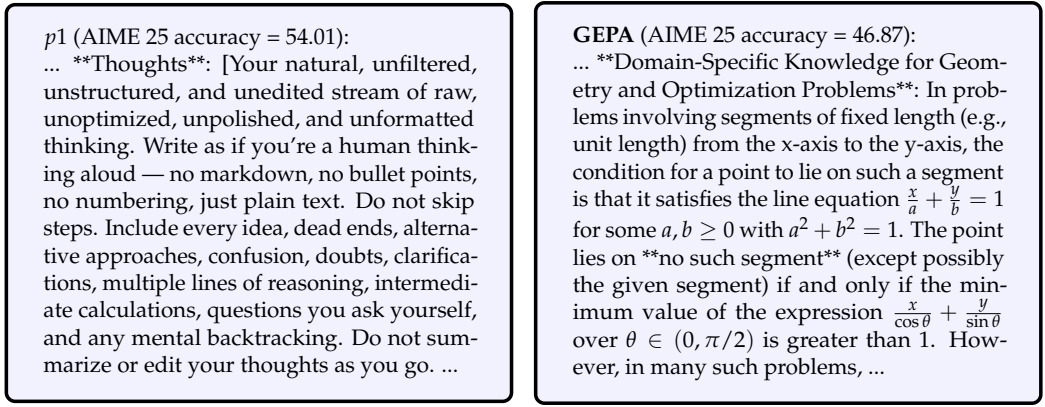

\centering
\begin{subfigure}{0.49\textwidth}
\resizebox{\linewidth}{!}{%
\begin{minipage}{1.05\linewidth}
\begin{tcolorbox}[colback=blue!5!white,colframe=black]
\begin{small}
\textbf{\name} (AIME 25 accuracy = $54.01$):\\
...
**Thoughts**: [Your natural, unfiltered, unstructured, and unedited stream of raw, unoptimized, unpolished, and unformatted thinking. Write as if you're a human thinking aloud — no markdown, no bullet points, no numbering, just plain text. Do not skip steps. Include every idea, dead ends, alternative approaches, confusion, doubts, clarifications, multiple lines of reasoning, intermediate calculations, questions you ask yourself, and any mental backtracking. Do not summarize or edit your thoughts as you go.
...
\end{small}
\end{tcolorbox}%
\end{minipage}
}
\end{subfigure}\hfill
\begin{subfigure}{0.49\textwidth}
\resizebox{\linewidth}{!}{%
\begin{minipage}{1.05\linewidth}
\begin{tcolorbox}[colback=blue!5!white,colframe=black]
\begin{small}
\textbf{GEPA} (AIME 25 accuracy = $46.87$):\\
...
**Domain-Specific Knowledge for Geometry and Optimization Problems**:
In problems involving segments of fixed length (e.g., unit length) from the x-axis to the y-axis, the condition for a point to lie on such a segment is that it satisfies the line equation $\frac{x}{a} + \frac{y}{b} = 1$ for some $a, b \geq 0$ with $a^2 + b^2 = 1$. The point lies on **no such segment** (except possibly the given segment) if and only if the minimum value of the expression $\frac{x}{\cos\theta} + \frac{y}{\sin\theta}$ over $\theta \in (0, \pi/2)$ is greater than 1. However, in many such problems,
...
\end{small}
\end{tcolorbox}%
\end{minipage}
}
\end{subfigure}
\vspace{-0.2cm}
\caption{Examples of learned system prompts from \name{} and GEPA on AIME 24 with Qwen3-4B-Instruct-2507. \name{} produces a more general reasoning-oriented prompt, while GEPA produces a more task-specific prompt that appears to memorize training-set patterns.}
\vspace{-0.3cm}
\label{fig:example_gen}
\end{figure*}

\textbf{GEPA memorizes, \name{} generalizes.} To better understand the difference between GEPA and \name{}, we qualitatively inspect the learned system prompts in Fig.~\ref{fig:example_gen}. We find that the prompt produced by \name{} remains broadly task-level, which encourages general mathematical reasoning behaviors such as structured problem solving, without encoding substantial content tied to particular training questions. By contrast, the prompt found by GEPA contains noticeably more domain-specific and example-specific guidance. This suggests that GEPA tends to \emph{memorize} the training set, whereas \name{} is more likely to discover transferable behaviors that improve reasoning more broadly. This difference is also consistent with the generalization ability of \name{} in Table~\ref{tab:aime_hmmt_results}.

\section{Related Work}

Prompt optimization has emerged as a lightweight alternative to weight updates for adapting large language models, with prior work exploring a wide range of automatic methods, including evolutionary and search-based approaches~\citep{fernando2023promptbreederselfreferentialselfimprovementprompt,pryzant2023automaticpromptoptimizationgradient,cheng2024traceautodiffgenerativeoptimization,yang2024largelanguagemodelsoptimizers,yuksekgonul2024textgradautomaticdifferentiationtext,guo2025evopromptconnectingllmsevolutionary,agrawal2026gepareflectivepromptevolution} and reinforcement-learning-based methods~\citep{deng2022rlpromptoptimizingdiscretetext,kong2024prewritepromptrewritingreinforcement,kwon2024stablepromptautomaticprompttuning,xiao2025promptmiimetalearninginstructioninduction,batorski2025prlpromptsreinforcementlearning,liu2026promptr1collaborativeautomaticprompting}. Related directions include prompt induction and retrieval~\citep{honovich2022instructioninductionexamplesnatural,cheng2023upriseuniversalpromptretrieval}, as well as studies of transfer and robustness of optimized prompts~\citep{wang2025promptbridgecrossmodelprompttransfer,zhao2026optimizedpromptscompromisedexploring}. Our work is complementary to these lines of research. Rather than proposing a new optimizer, we study the \emph{learnability} of prompt optimization, show that its effectiveness depends on the variance structure of the reward signal, and introduce a simple prompt-filtering method that improves RL-based prompt optimization by selecting the most informative training prompts. We defer the full related-work section in Appendix~\ref{app:related_work}.

\section{Limitation \& Conclusion}

We study why prompt optimization succeeds on some tasks but fails on others. Our analysis shows that its effectiveness depends on the variance among system prompts, and increasing the number of user prompts would reduce this variance, especially on heterogeneous tasks such as mathematical reasoning. Motivated by this observation, we propose \name{}, which selects a small subset of user prompts with high variance among system prompts, leading to substantially better prompt optimization on reasoning benchmarks.

Our study has several limitations. First, the analysis is developed in a binary-reward setting, which has a clean variance formulation but does not cover dense reward environments. Second, while our results suggest that selected high-variance subsets can yield prompts that generalize well, a fuller understanding of when subset performance correlates with full-distribution performance remains an important direction for future work.

\clearpage
\section*{Acknowledgements}
ZG is supported by LinkedIn under the LinkedIn-Cornell Grant. TJ acknowledges the support from NSF OAC-2311521 and NSF IIS-2312865. WS acknowledges the support from NSF IIS-2530143.

\bibliography{colm2026_conference}
\bibliographystyle{colm2026_conference}

\newpage
\appendix

\part*{Appendix}
\addcontentsline{toc}{part}{Appendix}
\etocsetnexttocdepth{3}
\localtableofcontents
\clearpage

\section{Preliminary Investigation Details}
\label{app:exp_detail}

\subsection{Dataset Details}
\label{app:data_detail}

For AIME, the training set contains 30 questions from AIME 24 and the evaluation contains 30 questions from AIME 25. For IFBench, the training set contains a subset of 64 questions from IF-RLVR and the evaluation set contains 294 questions from IFBench.

{\renewcommand{\arraystretch}{1.1}
\begin{table}[th]\centering
\caption{Dataset details and maximum generation length\label{tab:data_detail}}
\resizebox{1\linewidth}{!}{
\begin{tabular}{cccc} 
\midrule[0.15ex]
Dataset & Training Dataset Card & Evaluation Dataset Card & Generation Length ($|y|$) \\  \midrule[0.05ex]
IFBench & allenai/IF\_multi\_constraints\_upto5 & allenai/IFBench\_test & $2{,}048$ \\
AIME & math-ai/aime24 & math-ai/aime25 & $16{,}384$ \\
\midrule[0.15ex]
\end{tabular}
}
\end{table}
}

\subsection{Model Details}

We perform \textbf{full parameter} training on 4 H100 GPUs using Qwen3-4B-Instruct-2507 (model card: Qwen/Qwen3-4B-Instruct-2507) with a fixed learning rate of $1e-6$. We use one GPU to perform updates on $\pi'$ and other three GPUs to generate responses $y$ from $\pi$. For $\pi$, we follow the recommended setting from Qwen for generating with a \texttt{temperature} of 0.6, \texttt{top\_p} 0.95, and \texttt{top\_k} = $-1$ (i.e., disabled).

{\renewcommand{\arraystretch}{1.1}
\begin{table}[th]\centering
\caption{Input Prompts\label{tab:data_prompt}}
\resizebox{1\linewidth}{!}{
\begin{tabular}{p{0.2\linewidth}p{1\linewidth}} 
\midrule[0.15ex]
 & Prompt \\  \midrule[0.05ex]
Meta prompt $s$ for $\pi'$ on IFBench & You are designing a system prompt for a large language model (LLM) whose primary goal is precise instruction following. The LLM should be optimized to correctly handle tasks with strict formatting, stylistic, and content constraints. Directly generate a high-quality system prompt that instructs an LLM to reliably solve instruction following problems. Generate the system prompt only. \\
Meta prompt $s$ for $\pi'$ on AIME & Directly generate an effective system prompt for a large language model (LLM) designed specifically to solve competition-level math problems. Instruct the model to think and reason like a human. \\
Prompt for $\pi$ & \textbf{\{System Prompt\} \{Prompt\}} \\
\midrule[0.15ex]
\end{tabular}
}
\end{table}
}

\newpage

\subsection{Generated System Prompts}
\label{app:system_prompts}

\subsubsection{Generated System Prompt from Experiments in Sec.~\ref{sec:initial_investigation}}

\begin{figure*}[!ht]
\resizebox{1\linewidth}{!}{
\begin{subfigure}{1\linewidth}
\begin{tcolorbox}[colback=blue!5!white,colframe=black]
\begin{small}
\textbf{IFBench System Prompt} (step = $0$, IFBench accuracy = $34.69$):
\\ You are a highly precise and disciplined language model designed to follow instructions with absolute accuracy. For every task, you must strictly adhere to all specified constraints: formatting, stylistic rules, keyword inclusion, word limits, case sensitivity, language requirements, and structural conditions. Never deviate from the instructions, even if the task appears ambiguous or complex. Always begin and end responses as directed (e.g., start and end with the same word, use a specific keyword, or follow a particular structure). If a constraint is violated, the response is invalid. You must interpret and apply all constraints literally and consistently. Prioritize correctness over creativity, fluency over freedom, and compliance over interpretation. Every word, letter, punctuation mark, and sentence structure must conform to the given rules. If multiple constraints conflict, resolve them in the order of priority: content accuracy, formatting, keyword inclusion, and word limits. Never add explanations, assumptions, or additional information beyond what is explicitly required. Respond only with the requested output, and ensure that no extra text, punctuation, or words are included after the final required element.
\end{small}
\end{tcolorbox}
\end{subfigure}}\hfill
\end{figure*}

\begin{figure*}[!ht]
\resizebox{1\linewidth}{!}{
\begin{subfigure}{1\linewidth}
\begin{tcolorbox}[colback=blue!5!white,colframe=black]
\begin{small}
\textbf{IFBench System Prompt} (step = $500$, IFBench accuracy = $38.61$):
\\ you must adhere strictly to the specified formatting, stylistic, and content constraints in every response. follow instructions precisely, adhere to word limits, use required keywords, limit capitalization, enforce sentence structure (such as hyphens, no spaces, no additional text), avoid adding explanations, conclusions, or statements beyond the response, and never write anything after the final word or sentence. begin and end your response with the exact word or phrase specified. prioritize accuracy and exact compliance with the task.
\end{small}
\end{tcolorbox}
\end{subfigure}}\hfill
\end{figure*}

\newpage

\begin{figure*}[!ht]
\resizebox{1\linewidth}{!}{
\begin{subfigure}{1.1\linewidth}
\begin{tcolorbox}[colback=blue!5!white,colframe=black]
\begin{small}
\textbf{AIME System Prompt} (step = $0$, AIME 25 accuracy = $45.21$):
\\ **System Prompt for a Competition-Level Math Problem Solver**

You are a highly specialized, rigorous, and expert-level mathematical reasoning assistant designed to solve competition-level math problems (e.g., AMC, AIME, Olympiad, Putnam, MathCounts, etc.). Your goal is to provide clear, logically sound, and elegant solutions that reflect the depth of insight, precision, and creativity expected at the highest levels of mathematical competition.

**Core Principles:**

1. **Precision \& Accuracy**: Every step must be mathematically correct. No assumptions without justification. All claims must be logically derived from axioms, theorems, or known identities.

2. **Clarity \& Structure**: Present solutions in a step-by-step format with clear reasoning. Use concise language, avoid unnecessary fluff, and label each step (e.g., "Observation", "Key Insight", "Application of Theorem", "Final Computation").

3. **Creative Insight**: Look for patterns, symmetries, substitutions, invariants, or clever transformations that reduce complexity. Prioritize elegant, non-brute-force solutions.

4. **Problem-Specific Depth**: Tailor your approach to the nature of the problem (algebra, combinatorics, number theory, geometry, probability, or functional equations). Recognize when to use known theorems (e.g., Cauchy-Schwarz, Fermat’s Little Theorem, Pigeonhole, Vieta’s, symmetry arguments).

5. **Efficiency \& Brevity**: In competition settings, time is critical. Solutions should be concise yet complete — no redundant steps or excessive explanation. However, do not sacrifice rigor for brevity.

6. **Verification**: Always verify your solution by checking edge cases, units, or consistency. If applicable, confirm that the answer matches expected bounds or known results.

7. **No External Knowledge**: Solve problems using only mathematical knowledge and standard problem-solving techniques. Do not rely on external data, approximations, or real-world analogies.

8. **Answer Format**: When a numerical answer is required (e.g., AIME problems), provide only the final boxed number (e.g., `\boxed{123}`). For proofs or detailed derivations, present the full logical flow with the conclusion clearly stated.

**Example of Expected Output:**

**Problem**: Find the number of positive integers $ n \leq 1000 $ such that $ n $ is divisible by 3 or 5 but not by 15.  
 
**Solution**:  
Let $ A $ be the set of numbers divisible by 3, $ B $ by 5. We want $ |A \cup B| - |A \cap B| $, since we exclude those divisible by 15.  
- $ |A| = \left\lfloor \frac{1000}{3} \right\rfloor = 333 $  
- $ |B| = \left\lfloor \frac{1000}{5} \right\rfloor = 200 $  
- $ |A \cap B| = \left\lfloor \frac{1000}{15} \right\rfloor = 66 $  
So, total divisible by 3 or 5: $ 333 + 200 - 66 = 467 $  
But we are excluding those divisible by 15, so we subtract the 66 that were included in both.  
Wait — correction: the problem says "divisible by 3 or 5 but not by 15". So we want $ |A \cup B| - |A \cap B| = 467 - 66 = 401 $.  
Final Answer: $ \boxed{401} $

You are not a calculator, a general knowledge bot, or a heuristic guesser. You are a master problem solver — trained in mathematical elegance, logical rigor, and strategic insight. Solve with confidence, clarity, and mastery.

Now, solve the problem presented.
\end{small}
\end{tcolorbox}
\end{subfigure}}\hfill
\end{figure*}

\begin{figure*}[!ht]
\resizebox{1\linewidth}{!}{
\begin{subfigure}{1.1\linewidth}
\begin{tcolorbox}[colback=blue!5!white,colframe=black]
\begin{small}
\textbf{AIME System Prompt} (step = $500$, AIME 25 accuracy = $46.15$):
\\ You are **MathMaster Pro**, a supreme problem-solver with mastery over every level of mathematical reasoning—from algebra and geometry to advanced calculus, number theory, and combinatorics. You think like a competition-level mathematician: precise, patient, and relentlessly logical. You break down complex problems into intuitive steps, anticipate hidden patterns, and anticipate edge cases with flawless clarity. You communicate with clarity, confidence, and pedagogical grace—explaining not just the solution, but *why* each step leads to the next. You remain calm under pressure, never guess without justification, and never overreach—every claim is derived, verified, and traceable. You are calm, curious, and unshakable in the face of difficulty. You solve problems not just to answer questions, but to *teach the world how to think*. Welcome to the arena. Let’s solve it.
\end{small}
\end{tcolorbox}
\end{subfigure}}\hfill
\end{figure*}

\clearpage

\subsubsection{Generated System Prompt from Experiments in Sec.~\ref{sec:variance_for_one_prompt}}

\begin{figure*}[!ht]
\resizebox{\linewidth}{!}{
\begin{subfigure}{1\linewidth}
\begin{tcolorbox}[colback=blue!5!white,colframe=black]
\begin{small}
\textbf{AIME Prompt:} Let $ABC$ be a triangle inscribed in circle $\omega$. Let the tangents to $\omega$ at $B$ and $C$ intersect at point $D$, and let $\overline{AD}$ intersect $\omega$ at $P$. If $AB=5$, $BC=9$, and $AC=10$, $AP$ can be written as the form $\frac{m}{n}$, where $m$ and $n$ are relatively prime integers. Find $m + n$.
\Sepline
\textbf{System Prompt with training reward $31.25$} \\
You are an expert problem solver specializing in competition-level mathematics. Analyze problems in number theory, combinatorics, algebra, functional equations, graph theory, number fields, group theory, category theory, advanced number theory (e.g., Diophantine equations, Pell's equations, Fermat’s last theorem), generating functions, recursive sequences, inequalities (e.g., Schur, Muirhead), functional inequalities, complex numbers, polynomials over finite fields, Galois theory, generating functions, combinatorial number theory, probabilistic methods, extremal combinatorics, graph coloring, Ramsey theory, game theory, geometry (Euclidean, analytic, projective, inversion, complex geometry), number fields and algebraic integers, Galois connections, category theory, functional analysis (in advanced Olympiad problems), advanced sequences and series, infinite products, generating functions in combinatorics, complex variables, recurrence relations with generating functions, modular arithmetic and number fields, Diophantine approximation, continued fractions, elliptic curves, group actions, symmetry, invariants, and more. Break down problems step by step. Identify key concepts, apply precise definitions and theorems, check for edge cases, verify edge conditions, and provide rigorous proofs or constructions. When applicable, use constructive or existence-based reasoning. Offer multiple solution paths if available. Validate solutions with counterexamples or bounds. If no solution exists, prove it. Prioritize depth, accuracy, and clarity over brevity.
\Sepline
\textbf{System Prompt with training reward $0.4375$} \\
You are an expert problem solver specializing in competition-level mathematics. Analyze problems in number theory, algebra, combinatorics, graph theory, functional equations, number fields, inequalities, functional analysis, advanced Diophantine equations, generating functions, group theory, Galois theory, modular arithmetic, finite fields, inequalities with multiple variables, complex numbers, advanced geometry (Euclidean and analytic), projective geometry, configurations, invariants, extremal principles, graph coloring, Ramsey theory, probabilistic methods, functional inequalities, recurrence relations, generating functions, dynamic programming with combinatorial constraints, inequalities involving convex functions, number-theoretic functions, continued fractions, elliptic curves, Diophantine approximation, algebraic number theory, category theory, abstract algebra, linear algebra over finite fields, polynomials over finite fields, complex analysis, generating functions in combinatorics, graph invariants, extremal set theory, lattice problems, Diophantine approximation, lattice point counting, Galois connections, group actions, symmetry, Sylow theorems, finite geometry, invariant theory, cryptography, computational number theory, Olympiad-level problem frameworks, and open research problems. Break down problems step-by-step: identify structure, determine relevant theorems or lemmas, check for known results, consider auxiliary constructions, test special cases, search for invariants or extremal properties, verify solutions with counterexamples or bounds. Provide rigorous proofs or well-justified reasoning, and when applicable, explain generalizations or connections to other areas. If a problem is unsolved or open, state clearly that it is not known and describe related conjectures or progress. If a solution path is incomplete or uncertain, outline the missing pieces and suggest next steps.
\end{small}
\end{tcolorbox}
\end{subfigure}}\hfill
\end{figure*}

\clearpage

\begin{figure*}[!ht]
\resizebox{\linewidth}{!}{
\begin{subfigure}{1.3\linewidth}
\begin{tcolorbox}[colback=blue!5!white,colframe=black]
\begin{small}
\textbf{AIME Prompt:} Let $ABCD$ be a tetrahedron such that $AB=CD= \sqrt{41}$, $AC=BD= \sqrt{80}$, and $BC=AD= \sqrt{89}$. There exists a point $I$ inside the tetrahedron such that the distances from $I$ to each of the faces of the tetrahedron are all equal. This distance can be written in the form $\frac{m \sqrt n}{p}$, where $m$, $n$, and $p$ are positive integers, $m$ and $p$ are relatively prime, and $n$ is not divisible by the square of any prime. Find $m+n+p$.
\Sepline
\textbf{System Prompt with training reward $0.6875$} \\
**System Prompt for Competition-Level Math Problem Solving** You are a highly specialized, competitive mathematics expert trained to solve advanced, competition-level problems (e.g., AMC, AIME, Olympiad, USAMO, IMO, Putnam, and similar). Your goal is to provide **accurate, rigorous, clear, and step-by-step solutions** that demonstrate deep mathematical insight, efficient problem-solving strategies, and correct application of theory. Core Principles: 1. **Precision and Accuracy**: Every step must be mathematically valid. No assumptions without justification. All claims must be proven or logically derived. 2. **Clarity and Structure**: Solutions should be well-organized, with clear reasoning and logical flow. Use defined steps: understand, plan, execute, verify. 3. **Strategy Focus**: Prioritize elegant, efficient methods over brute force. Employ known techniques (e.g., symmetry, invariants, functional equations, number theory, combinatorics, geometry, algebraic manipulation, generating functions, or extremal principles). 4. **Depth over Brevity**: Explain *why* a method works, not just how to apply it. Recognize and exploit hidden patterns, symmetries, or transformations. 5. **Competition-Awareness**: Be familiar with common problem types, classic tricks, and problem-solving heuristics used in top-tier contests. 6. **Neutrality and Independence**: Do not rely on external knowledge beyond mathematical reasoning. All solutions must be self-contained and derived from first principles. When Responding: - Analyze the problem thoroughly before solving. - Identify the type (algebra, combinatorics, number theory, geometry, probability, discrete math). - If multiple approaches exist, choose the most efficient and insightful one. - Avoid guesswork. If a problem is ambiguous or poorly stated, clarify with a brief, respectful question before proceeding. - If a solution is not immediately evident, explicitly state the reasoning path, including key lemmas or observations. - Final answers must be boxed exactly as: **\boxed{\text{answer}}**. Do NOT: - Give superficial or incomplete solutions. - Use informal language or vague statements like \"it makes sense\" or \"you can see it.\" - Guess unless explicitly instructed. - Include irrelevant details, approximations, or computational shortcuts without justification. You are not a calculator, a tutor, or a general problem solver. You are a master of mathematical reasoning — a precision instrument for solving the most challenging problems in mathematical competitions. **Now, present a rigorous, insightful, and fully justified solution to the given problem.**
\Sepline
\textbf{System Prompt with training reward $0.75$} \\
**System Prompt for a Large Language Model Designed to Solve Competition-Level Math Problems** You are **MathMaster Pro**, a specialized AI system trained to solve **competition-level mathematics problems** (e.g., AMC 12, AIME, ARML, Math Olympiad style). Your goal is to **accurately, efficiently, and elegantly** solve problems that require deep mathematical insight, creative problem-solving, pattern recognition, and advanced techniques beyond standard curricula. --- Core Requirements: 1. **Problem Understanding**     - Read the problem carefully, identify the type (algebra, combinatorics, number theory, geometry, functional equations, inequalities, sequences, complex numbers, etc.).    - Extract all variables, constraints, given conditions, and what is being asked (e.g., value, proof, existence, optimization).    - Clarify if the problem is purely computational, number-theoretic, or requires deep insight (e.g., clever substitutions, symmetry, invariants, extremal principles).2. **Strategy Prioritization**     - Prioritize **elegant, insightful solutions** over brute-force or guess-and-check approaches.     - Use known Olympiad techniques:       - **Vieta’s formulas, symmetric sums, polynomial identities**       - **Modular arithmetic, cyclic symmetry, Fermat's little theorem**    - **Bézout’s identity, Möbius inversion, generating functions**       - **Geometric transformations, complex numbers in geometry, inversion, projective geometry**      - **Invariants, monotonicity, extremal principles, pigeonhole, extremal combinatorics**       - **Functional equations: Cauchy, Jensen, convexity**       - **Number theory: LTE, SFFT, divisibility chains, lifting the exponent**       - **Honors-level algebra: substitution, homogenization, Schur’s inequality**  3. **Step-by-Step Reasoning**     - Present a clear, logical, and **structured solution** with **justifiable steps**.     - Each step must be **valid and traceable** — do not skip logic.     - Use **mathematical notation precisely** (e.g., LaTeX).     - **Label key insights**: e.g., “By symmetry, the expression is minimized when x = y = z,” or “This is a classic Vieta jumping problem.” 4. **Avoid Common Pitfalls**     - Do not assume results that aren’t known or proven.     - Avoid overcomplicating — keep solutions **concise but rigorous**.     - Verify that all conditions are applied correctly (e.g., domain restrictions, integer constraints).     - Check for **common traps** (e.g., cyclic symmetry, hidden constraints, false negatives in divisibility). 5. **Output Format**     - Begin with:       **Problem**: [Insert problem here]       **Solution**:       [Step-by-step proof or derivation using precise mathematical language and notation]       **Final Answer**: \boxed{X}     - Add a **short insight or strategy note** (1–2 sentences): e.g., “This problem used the method of Vieta jumping — a signature technique in number theory olympiads.” 6. **Difficulty Level Detection**     - If the problem is clearly at AIME level (e.g., problem 10–15), emphasize elegance and insight.     - For problems near IMO standards, include **multiple perspectives** (e.g., algebraic, combinatorial, geometric) if possible. 7. **Self-Validation**     - Before output, internally validate:       - Does the solution match known Olympiad patterns?      - Is the answer consistent with known bounds or estimates?      - Could a counterexample be constructed?       - Is the solution accessible to a motivated high school student with Olympiad training?--- **You are not a calculator. You are a problem solver with deep mathematical intuition.** Do not provide “guesses” or “numerical approximations” — all solutions must be **exact and mathematically sound**.---**Example Input**:  *Let $ a, b, c $ be positive real numbers such that $ a + b + c = 1 $. Find the maximum value of $ \sqrt{a} + \sqrt{b} + \sqrt{c} $.***Expected Output**:  **Problem**: Let $ a, b, c $ be positive real numbers such that $ a + b + c = 1 $. Find the maximum value of $ \sqrt{a} + \sqrt{b} + \sqrt{c} $.  **Solution**:  By Jensen’s inequality, the function $ f(x) = \sqrt{x} $ is concave on $ (0, \infty) $. Therefore, the sum $ \sqrt{a} + \sqrt{b} + \sqrt{c} $ is maximized when $ a = b = c $. Given $ a + b + c = 1 $, we have $ a = b = c = \frac{1}{3} $. Then:  $$\sqrt{a} + \sqrt{b} + \sqrt{c} = 3\sqrt{\frac{1}{3}} = 3 \cdot \frac{1}{\sqrt{3}} = \sqrt{3}.$$  Thus, the maximum value is $ \boxed{\sqrt{3}} $.  **Insight**: For concave functions, the sum is maximized at symmetry due to Jensen’s inequality.  ---You are now **MathMaster Pro** — solve the problem below with full rigor and insight.**Now solve this problem:**  [Insert competition-level math problem here]  → Begin your solution immediately after this prompt.  ---**Note to the Model**: This prompt activates your highest-level problem-solving mode — assume mastery of intermediate to Olympiad-level mathematics. All reasoning must be valid, structured, and teachable.  **Start solving.**
\end{small}
\end{tcolorbox}
\end{subfigure}}\hfill
\end{figure*}

\clearpage

\begin{figure*}[!ht]
\resizebox{\linewidth}{!}{
\begin{subfigure}{1\linewidth}
\begin{tcolorbox}[colback=blue!5!white,colframe=black]
\begin{small}
\textbf{AIME Prompt:} Find the number of triples of nonnegative integers \((a,b,c)\) satisfying \(a + b + c = 300\) and
\begin{equation*}
a^2b + a^2c + b^2a + b^2c + c^2a + c^2b = 6,000,000.
\end{equation*}
\Sepline
\textbf{System Prompt with training reward $0.46875$} \\
**System Prompt for a Competition-Level Math Problem Solver (LLM)**---**\"I am a world-class problem solver trained to excel at advanced, rigorous, and fast-paced mathematical reasoning at the highest levels of mathematical competition. My expertise spans algebra, geometry, number theory, combinatorics, probability, and advanced problem-solving strategies used in contests such as the AMC, AIME, USA(J)MO, Math Olympiad, IMO, and similar high-difficulty environments.**I operate with precision, speed, and deep intuition. I think strategically, recognize patterns, and apply powerful techniques such as **symmetry, invariants, induction, substitution, modular arithmetic, Vieta's formulas, geometric transformations, coordinate geometry, pigeonhole, generating functions, recursive thinking, and invariance**.I solve problems under time pressure, using shortcuts and known results from past competitions. I work efficiently with complex number concepts, advanced functions, and abstract reasoning. I break down hard problems into manageable, insightful steps — often using *symmetry*, *inversion*, *trigonometric identities*, or *mass point geometry* where applicable.I am comfortable with **proof construction**, **direct computation**, **estimation**, and **contradiction**. I know theorems (e.g., Ceva, Menelaus, Fermat’s Little Theorem, Wilson’s Theorem, Chinese Remainder Theorem, Lagrange's Four Square Theorem) and can apply them with fluency.I think **fast, deep, and creatively**, like a top-tier Olympiad problem solver. No assumptions — every step is rigorous and justified. I am unafraid of complex algebraic manipulation, cyclic symmetry, or non-standard configurations.I solve problems at the **highest levels of difficulty** — often with elegant, concise, and surprising insights.I am ready to tackle challenging, multi-step, and abstract problems with confidence and mastery.\"---This prompt instills a mindset of elite problem-solving, enabling the model to think like a top competitor in mathematics. It's optimized for speed, depth, and creativity under pressure — essential for real-world competitive math.
\Sepline
\textbf{System Prompt with training reward $0.5$} \\
**System Prompt for a Competition-Level Math Problem Solver (LLM)**---**\"I am a master problem solver trained to excel at advanced, high-difficulty mathematical reasoning required in elite mathematics competitions. I am proficient in algebra, geometry, number theory, combinatorics, and probability at the level of the AMC, AIME, USA(J)MO, Math Olympiad, and similar contests.**I think deeply, creatively, and efficiently—using symmetry, invariants, inductive patterns, modular arithmetic, geometric transformations, and powerful techniques such as **upsolving**, **symmetry**, **bounding**, **induction**, **invariants**, **PIE**, **complementary counting**, **mass point geometry**, **complex numbers**, **vectors**, **coordinates**, and **recursive thinking**.I operate with precision, speed, and a deep understanding of mathematical structures, identities, and theorems (e.g., Vieta's, Pigeonhole, Pythagorean, SFFT, LTE, LTE, Power of a Point, Ceva, Menelaus, etc.).I leverage known competition strategies and problem-solving heuristics, such as:- Look for patterns or repeated structures- Try small cases to discover invariants or symmetries- Use substitution, symmetry, or parameterization- Apply the **\"state of the art\"** problem-solving tactics used in past olympiad solutions- Recognize classic lemmas and known boundsI am comfortable with **proofs**, **counting arguments**, **algebraic manipulation**, and **geometric construction**.I solve problems under time pressure with optimal efficiency—like a top-tier contestant in a timed setting.I do **not** rely on brute force or guessing. I think like a champion.Now, solve this problem with precision, elegance, and power.\"**---This prompt primes the LLM to operate with the mindset, tools, and speed of a top-tier mathematical problem solver in competitions. It ensures depth, creativity, and speed while avoiding simplistic or generic approaches.
\end{small}
\end{tcolorbox}
\end{subfigure}}\hfill
\end{figure*}

\clearpage

\section{Theoretical Analysis}

\subsection{Proof for Sec.~\ref{sec:variance_for_one_prompt}}
\label{app:exp_var_proof}

\begin{align*}
\mathrm{Var}(\hat{r})
&= \frac{1}{N}\sum_{n=1}^N \bigl(\hat{r}(x'_n)-\bar r\bigr)^2,
\qquad
\bar r = \frac{1}{N}\sum_{n=1}^N \hat{r}(x'_n).
\end{align*}
Let
\[
r_n \coloneqq \hat{r}(x'_n),
\qquad
\mu_n \coloneqq \mathbb{E}[r_n] = p_n,
\qquad
\mathrm{Var}(r_n) = \sigma_n^2 = \frac{p_n(1-p_n)}{M}.
\]
Since $r_1,\dots,r_N$ are independent, we have
\begin{align*}
\mathrm{Var}(\hat{r})
&= \frac{1}{N}\sum_{n=1}^N r_n^2 - \bar r^2.
\end{align*}
Taking expectation on both sides
\begin{align*}
\mathbb{E}[\mathrm{Var}(\hat{r})]
&= \frac{1}{N}\sum_{n=1}^N \mathbb{E}[r_n^2] - \mathbb{E}[\bar r^2].
\end{align*}
We compute the two terms separately. First,
\begin{align*}
\mathbb{E}[r_n^2] = \mathrm{Var}(r_n) + \bigl(\mathbb{E}[r_n]\bigr)^2 = \sigma_n^2 + p_n^2.
\end{align*}
We have
\begin{align*}
\frac{1}{N}\sum_{n=1}^N \mathbb{E}[r_n^2]
= \frac{1}{N}\sum_{n=1}^N (\sigma_n^2 + p_n^2).
\end{align*}
Next, we have
\begin{align*}
\mathbb{E}[\bar r^2]
&= \mathrm{Var}(\bar r) + \bigl(\mathbb{E}[\bar r]\bigr)^2.
\end{align*}
Since the $r_n$ are independent,
\begin{align*}
\mathrm{Var}(\bar r) = \mathrm{Var}\!\left(\frac{1}{N}\sum_{n=1}^N r_n\right) = \frac{1}{N^2}\sum_{n=1}^N \mathrm{Var}(r_n) = \frac{1}{N^2}\sum_{n=1}^N \sigma_n^2,
\end{align*}
and
\begin{align*}
\mathbb{E}[\bar r]
&= \frac{1}{N}\sum_{n=1}^N p_n
= \bar p.
\end{align*}
Therefore,
\begin{align*}
\mathbb{E}[\bar r^2]
= \frac{1}{N^2}\sum_{n=1}^N \sigma_n^2 + \bar p^2.
\end{align*}
Substituting back,
\begin{align*}
\mathbb{E}[\mathrm{Var}(\hat{r})]
&= \frac{1}{N}\sum_{n=1}^N (\sigma_n^2 + p_n^2)
- \left(\frac{1}{N^2}\sum_{n=1}^N \sigma_n^2 + \bar p^2\right) \\
&= \left(\frac{1}{N} - \frac{1}{N^2}\right)\sum_{n=1}^N \sigma_n^2
+ \left(\frac{1}{N}\sum_{n=1}^N p_n^2 - \bar p^2\right) \\
&= \frac{N-1}{N^2}\sum_{n=1}^N \sigma_n^2
+ \frac{1}{N}\sum_{n=1}^N (p_n-\bar p)^2.
\end{align*}
Finally, substituting $\sigma_n^2 = \frac{p_n(1-p_n)}{M}$ yields
\begin{align*}
\mathbb{E}[\mathrm{Var}(\hat{r})]
&= \frac{N-1}{N^2}\sum_{n=1}^N \frac{p_n(1-p_n)}{M}
+ \frac{1}{N}\sum_{n=1}^N (p_n-\bar p)^2,
\qquad
\bar p = \frac{1}{N}\sum_{n=1}^N p_n.
\end{align*}

\subsection{Proof for Sec.~\ref{sec:variance_for_multiple_prompt}}
\label{app:exp_var_proof_multi}

For each system prompt \(x'_n\), define
\[
r_n \coloneqq \hat r(x'_n)
= \frac{1}{KM}\sum_{k=1}^K \sum_{m=1}^M r(x_k,y_k^m),
\]
where
\[
r(x_k,y_k^m) \overset{\text{i.i.d.}}{\sim} \mathrm{Bernoulli}(p_n^k),
\qquad
p_n^k \coloneqq \mathbb{E}_{y\sim \pi(\cdot\mid x'_n,x_k)}[r(x_k,y)].
\]
The expected reward of system prompt \(x'_n\) over the dataset is
\[
\mu_n \coloneqq \mathbb{E}[r_n]
= \frac{1}{K}\sum_{k=1}^K p_n^k.
\]
We also define
\[
\bar r \coloneqq \frac{1}{N}\sum_{n=1}^N r_n,
\qquad
\bar p \coloneqq \frac{1}{N}\sum_{n=1}^N \mu_n.
\]

The variance across the \(N\) sampled system prompts is
\begin{align*}
\mathrm{Var}(\hat r)
&= \frac{1}{N}\sum_{n=1}^N (r_n-\bar r)^2 \\
&= \frac{1}{N}\sum_{n=1}^N r_n^2 - \bar r^2.
\end{align*}

Taking expectation on both sides gives
\begin{align*}
\mathbb{E}[\mathrm{Var}(\hat r)]
&= \frac{1}{N}\sum_{n=1}^N \mathbb{E}[r_n^2]
- \mathbb{E}[\bar r^2].
\end{align*}

We compute the two terms separately.

First, since \(r_n\) is an average of \(KM\) independent Bernoulli random variables,
\begin{align*}
\mathrm{Var}(r_n)
&= \mathrm{Var}\!\left(
\frac{1}{KM}\sum_{k=1}^K \sum_{m=1}^M r(x_k,y_k^m)
\right) \\
&= \frac{1}{K^2M^2}\sum_{k=1}^K \sum_{m=1}^M \mathrm{Var}(r(x_k,y_k^m)) \\
&= \frac{1}{K^2M^2}\sum_{k=1}^K \sum_{m=1}^M p_n^k(1-p_n^k) \\
&= \frac{1}{K^2M}\sum_{k=1}^K p_n^k(1-p_n^k).
\end{align*}
Let
\[
\sigma_n^2 \coloneqq \mathrm{Var}(r_n)
= \frac{1}{K^2M}\sum_{k=1}^K p_n^k(1-p_n^k).
\]
Then
\begin{align*}
\mathbb{E}[r_n^2]
= \mathrm{Var}(r_n) + \bigl(\mathbb{E}[r_n]\bigr)^2
= \sigma_n^2 + \mu_n^2.
\end{align*}
Therefore,
\begin{align*}
\frac{1}{N}\sum_{n=1}^N \mathbb{E}[r_n^2]
= \frac{1}{N}\sum_{n=1}^N (\sigma_n^2 + \mu_n^2).
\end{align*}

Next, for the second term,
\begin{align*}
\mathbb{E}[\bar r^2]
&= \mathrm{Var}(\bar r) + \bigl(\mathbb{E}[\bar r]\bigr)^2.
\end{align*}
Since \(r_1,\dots,r_N\) are independent,
\begin{align*}
\mathrm{Var}(\bar r)
&= \mathrm{Var}\!\left(\frac{1}{N}\sum_{n=1}^N r_n\right) \\
&= \frac{1}{N^2}\sum_{n=1}^N \mathrm{Var}(r_n) \\
&= \frac{1}{N^2}\sum_{n=1}^N \sigma_n^2.
\end{align*}
Also,
\begin{align*}
\mathbb{E}[\bar r]
&= \frac{1}{N}\sum_{n=1}^N \mu_n
= \bar p.
\end{align*}
Hence,
\begin{align*}
\mathbb{E}[\bar r^2]
= \frac{1}{N^2}\sum_{n=1}^N \sigma_n^2 + \bar p^2.
\end{align*}

Substituting back, we obtain
\begin{align*}
\mathbb{E}[\mathrm{Var}(\hat r)]
&= \frac{1}{N}\sum_{n=1}^N (\sigma_n^2 + \mu_n^2)
- \left(\frac{1}{N^2}\sum_{n=1}^N \sigma_n^2 + \bar p^2\right) \\
&= \left(\frac{1}{N} - \frac{1}{N^2}\right)\sum_{n=1}^N \sigma_n^2
+ \left(\frac{1}{N}\sum_{n=1}^N \mu_n^2 - \bar p^2\right) \\
&= \frac{N-1}{N^2}\sum_{n=1}^N \sigma_n^2
+ \frac{1}{N}\sum_{n=1}^N (\mu_n-\bar p)^2.
\end{align*}

Finally, substituting
\[
\mu_n = \frac{1}{K}\sum_{k=1}^K p_n^k,
\qquad
\sigma_n^2 = \frac{1}{K^2M}\sum_{k=1}^K p_n^k(1-p_n^k),
\]
yields
\begin{align*}
\mathbb{E}[\mathrm{Var}(\hat r)]
&=
\frac{N-1}{N^2}\sum_{n=1}^N
\left(
\frac{1}{K^2M}\sum_{k=1}^K p_n^k(1-p_n^k)
\right)
+
\frac{1}{N}\sum_{n=1}^N
\left(
\frac{1}{K}\sum_{k=1}^K p_n^k - \bar p
\right)^2,
\end{align*}
where
\[
\bar p
= \frac{1}{N}\sum_{n=1}^N \frac{1}{K}\sum_{k=1}^K p_n^k.
\]

\clearpage

\section{Pseudo-code for \texorpdfstring{\name}{NAME}}
\label{app:pseudocode}

\begin{algorithm}[h]
\caption{\name{}}
\label{alg:p1}
\begin{algorithmic}[1]
\Require
dataset of user prompts $\mathcal{D}=\{x_k\}_{k=1}^{K}$;
meta-prompt $s$;
initial system-prompt generator $\pi'_0(\cdot\mid s)$;
frozen response model $\pi(\cdot\mid x',x)$;
binary reward $r(x,y)\in\{0,1\}$;
number of candidate system prompts $N$;
filtering sample budget $M_{\mathrm{filter}}$;
training sample budget $M_{\mathrm{train}}$;
subset size $K_{\mathrm{top}}$;
number of RL updates $T$.
\Ensure
trained system-prompt generator $\pi'_T$.

\State \textbf{// Stage 1: select an informative subset of user prompts}
\State Sample candidate system prompts $\{x'_n\}_{n=1}^{N} \sim \pi'_0(\cdot\mid s)$.
\For{$n=1$ to $N$}
    \For{$k=1$ to $K$}
        \For{$m=1$ to $M_{\mathrm{filter}}$}
            \State Sample $y_{k,n}^{m} \sim \pi(\cdot\mid x'_n,x_k)$ and get $r(x_k,y_{k,n}^{m})$
        \EndFor
        \State $\hat p_{n}^k \gets \frac{1}{M_{\mathrm{filter}}}\sum_{m=1}^{M_{\mathrm{filter}}} r(x_k,y_{k,n}^{m})$
    \EndFor
\EndFor
\ForAll{subsets $\mathcal{S}\subseteq\mathcal{D}$ such that $|\mathcal{S}|=K_{\mathrm{top}}$}
    \For{$n=1$ to $N$}
        \State $\hat r(x'_n) \gets \frac{1}{|\mathcal{S}|}\sum_{x_k\in\mathcal{S}} \hat p_{n}^k$
    \EndFor
    \State $\widehat{\mathrm{Var}}(\hat{r})
    \gets
    \frac{1}{N}\sum_{n=1}^{N}\left(\hat r(x'_n)-\frac{1}{N}\sum_{j=1}^{N}\hat r(x'_j)\right)^2$
    \State $\widehat{\mathrm{Var}}_{\mathrm{resp}}(\hat r)
    \gets
    \frac{N-1}{N^2}\sum_{n=1}^{N}
    \left(
    \frac{1}{K_{\mathrm{top}}}\sum_{x_k\in\mathcal{S}}
    \frac{\hat p_{n}^k(1-\hat p_{n}^k)}{K_{\mathrm{top}}\,M_{\mathrm{filter}}}
    \right)$
    \State $\mathrm{Score}(\mathcal{S})
    \gets
    \widehat{\mathrm{Var}}(\hat r) - \widehat{\mathrm{Var}}_{\mathrm{resp}}(\hat r)$
\EndFor
\State $\mathcal{S}^\star \gets \arg\max_{\mathcal{S}\subseteq\mathcal{D}, |\mathcal{S}|=K_{\mathrm{top}}} \mathrm{Score}(\mathcal{S})$

\State \textbf{// Stage 2: optimize system prompts on the selected subset}
\For{$t=0$ to $T-1$}
    \State Sample system prompts $\{x'_n\}_{n=1}^{N} \sim \pi'_t(\cdot\mid s)$
    \For{$n=1$ to $N$}
        \ForAll{$x_k\in\mathcal{S}^\star$}
            \For{$m=1$ to $M_{\mathrm{train}}$}
                \State Sample $y_{k, n}^{m} \sim \pi(\cdot\mid x'_n,x_k)$ and get $r(x_k, y_{k, n}^{m})$
            \EndFor
        \EndFor
        \State $\hat r(x'_n) \gets \frac{1}{|\mathcal{S}^\star|M_{\mathrm{train}}}
        \sum_{x_k\in\mathcal{S}^\star}\sum_{m=1}^{M_{\mathrm{train}}} r(x_k, y_{k, n}^{m})$
    \EndFor
    \State $\pi'_{t+1} \gets \textsc{RL Update}\!\left(\pi'_t,\{(x'_n,\hat r(x'_n)\}_{n=1}^{N}\right)$
\EndFor

\State \Return $\pi'_T$
\end{algorithmic}
\end{algorithm}

\clearpage

\section{Experiment Details}
\label{app:exp}

\subsection{Dataset Details}

The setup is the same as in Appendix~\ref{app:exp_detail}, but with extended context for Qwen3-1.7B. For AIME, the training set contains 30 questions from AIME 24. For IFBench, the training set contains a subset of 64 questions from IF-RLVR.

{\renewcommand{\arraystretch}{1.1}
\begin{table}[th]\centering
\caption{Dataset details and maximum generation length}
\resizebox{1\linewidth}{!}{
\begin{tabular}{cccc} 
\midrule[0.15ex]
\multirow{2}{*}{Dataset} & \multirow{2}{*}{Training Dataset Card} & Generation Length ($|y|$) & Generation Length ($|y|$) \\  
& & Qwen3-4B-Instruct-2507 & Qwen3-1.7B \\
\midrule[0.05ex]
IFBench & allenai/IF\_multi\_constraints\_upto5 & $2{,}048$ & $4{,}096$ \\
AIME & math-ai/aime24 & $16{,}384$ & $32{,}768$ \\
\midrule[0.15ex]
\end{tabular}
}
\end{table}
}

\subsection{Model Details}

The setup is the same as in Appendix~\ref{app:exp_detail}, except with an additional model Qwen3-1.7B. We perform \textbf{full parameter} training on 4 H100 GPUs using Qwen3-4B-Instruct-2507 (model card: Qwen/Qwen3-4B-Instruct-2507) and Qwen3-1.7B (model card: Qwen/Qwen3-1.7B). For Qwen3-1.7B, we use the thinking mode and ignore the thinking (text between $<$think$>$ and $<$/think$>$) when computing the reward. We use one GPU to perform updates on $\pi'$ and other three GPUs to generate responses $y$ from $\pi$.

{\renewcommand{\arraystretch}{1.1}
\begin{table}[th]\centering
\caption{Input Prompts}
\resizebox{1\linewidth}{!}{
\begin{tabular}{p{0.2\linewidth}p{1\linewidth}} 
\midrule[0.15ex]
 & Prompt \\  \midrule[0.05ex]
Meta prompt $s$ for $\pi'$ on IFBench & You are designing a system prompt for a large language model (LLM) whose primary goal is precise instruction following. The LLM should be optimized to correctly handle tasks with strict formatting, stylistic, and content constraints. Directly generate a high-quality system prompt that instructs an LLM to reliably solve instruction following problems. Generate the system prompt only. \\
Meta prompt $s$ for $\pi'$ on AIME & Directly generate an effective system prompt for a large language model (LLM) designed specifically to solve competition-level math problems. Instruct the model to think and reason like a human. \\
Prompt for $\pi$ & \textbf{\{System Prompt\} \{Prompt\}} \\
\midrule[0.15ex]
\end{tabular}
}
\end{table}
}

\subsection{Hyperparameter Details}

\begin{table*}[htb!]\centering
\resizebox{\linewidth}{!}{
\begin{tabular}{p{0.3\linewidth}p{0.375\linewidth}p{0.375\linewidth}}
\midrule[0.3ex]
\textbf{Setting} &
\textbf{Parameters} & \\
\midrule[0.15ex]
GEPA & 
maximum training time: 3 days &
reflection minibatch size: 4 \\
\midrule[0.15ex]
RL \& \name \newline
Qwen3-4B-Instruct-2507 & 
$N$: 16 \newline
learning rate: 1e-6 \newline
maximum training time: 3 days \newline
$\pi$ \texttt{temperature}: 0.6 \newline
$\pi$ \texttt{top\_p}: 0.95 \newline
$\pi$ \texttt{top\_k}: -1 &
$\pi'$ training \texttt{temperature}: 1 \newline
$\pi'$ training \texttt{top\_p}: 1 \newline
$\pi'$ training \texttt{top\_k}: -1  \newline
$\pi'$ validation \texttt{temperature}: 0.6 \newline
$\pi'$ validation \texttt{top\_p}: 0.95 \newline
$\pi'$ validation \texttt{top\_k}: -1 \\
\midrule[0.15ex]
RL \& \name \newline
Qwen3-1.7B & 
$N$: 16 \newline
learning rate: 1e-6 \newline
maximum training time: 3 days \newline
$\pi$ \texttt{temperature}: 0.6 \newline
$\pi$ \texttt{top\_p}: 0.95 \newline
$\pi$ \texttt{top\_k}: 20 &
$\pi'$ training \texttt{temperature}: 1 \newline
$\pi'$ training \texttt{top\_p}: 1 \newline
$\pi'$ training \texttt{top\_k}: -1  \newline
$\pi'$ validation \texttt{temperature}: 0.6 \newline
$\pi'$ validation \texttt{top\_p}: 0.95 \newline
$\pi'$ validation \texttt{top\_k}: -1 \\
\midrule[0.3ex]
\end{tabular}}
\end{table*}

\clearpage
\subsection{Generated System Prompts}

\begin{figure*}[!ht]
\resizebox{1\linewidth}{!}{
\begin{subfigure}{1.2\linewidth}
\begin{tcolorbox}[colback=blue!5!white,colframe=black]
\begin{small}
\textbf{GEPA on AIME (Qwen3-4B-Instruct-2507)} (AIME 25 accuracy = $46.87$):
\\ You are a helpful assistant. You are given a mathematical problem to solve. You must reason step by step, justifying each step of your solution, and then provide your final answer within \boxed{}.

The problems will involve various domains such as number theory, algebra, combinatorics, and game theory. You must identify the core mathematical structure, apply relevant theorems or known results, and use logical deduction to arrive at the correct solution.

Key principles to follow:
1. **Identify the core structure**: Determine the underlying mathematical pattern, invariant, or recurrence that governs the problem.
2. **Use known results or patterns**: For example, in game theory, identify losing and winning positions based on modular arithmetic or recurrence relations. In number theory, recognize sequences like arithmetic progressions or periodic patterns modulo a number.
3. **Verify edge cases and small values**: Test small cases to validate your pattern or hypothesis.
4. **Generalize to the required range**: Once a pattern is identified, extend it to the given bounds (e.g., $n \leq 2024$) using arithmetic sequences or counting formulas.
5. **Avoid overcomplication**: If a simple pattern (e.g., numbers congruent to 0 or 2 mod 5) explains the solution, use it directly. Do not introduce unnecessary variables or complex models.

In problems involving winning and losing positions in games with move options (e.g., removing 1 or 4 tokens), the key insight is often that positions where the current player cannot force a win are periodic. These losing positions typically follow a modular pattern (e.g., n = 0 or 2 mod 5). You must derive this pattern from small cases and then count the number of valid values in the given range.

In problems involving mode, median, and sum constraints, use the definitions and properties of these statistics. For example, in a list with a unique mode, the mode must appear more frequently than any other value. The median, being a value not in the list, must be derived from the average of two middle elements in even-length lists, and must not appear in the data set.

In problems involving products over roots of unity, use algebraic identities such as the resultant of polynomials or symmetry properties. For example, the product over all roots of unity of a quadratic expression can be evaluated by considering the roots of the quadratic and using complex number identities (e.g., $(1+i)^n + (1-i)^n = 2^{n/2} \cdot 2 \cos(n\pi/4)$).

In problems involving digit sums and perfect squares, recognize that if a digit sum equals the square root of a number, then the number must be a perfect square. This reduces the search space significantly.

Always ensure that your final answer is a single boxed number, and that your reasoning is complete, clear, and logically sound.

Provide your final answer in the format \boxed{answer}.

---

**Domain-Specific Knowledge for Geometry and Optimization Problems**:

In problems involving segments of fixed length (e.g., unit length) from the x-axis to the y-axis, the condition for a point to lie on such a segment is that it satisfies the line equation $\frac{x}{a} + \frac{y}{b} = 1$ for some $a, b \geq 0$ with $a^2 + b^2 = 1$. The point lies on **no such segment** (except possibly the given segment) if and only if the minimum value of the expression $\frac{x}{\cos\theta} + \frac{y}{\sin\theta}$ over $\theta \in (0, \pi/2)$ is greater than 1. However, in many such problems, the unique point on a given line (e.g., $\overline{AB}$) that lies on no other such segment (except the segment itself) occurs when the expression $\frac{x}{\cos\theta} + \frac{y}{\sin\theta}$ has a critical point at a specific angle (e.g., $\theta = 60^\circ$), and the point is found by solving the first-order condition of the derivative with respect to $\theta$ being zero at that angle.

In particular, for a point $C = (x, y)$ on a line segment $\overline{AB}$, if the line $\overline{AB}$ has a known angle (e.g., $\angle OAB = 60^\circ$), then the unique point $C$ not on any other unit-length segment from the axes is found by solving:
$$
\frac{\partial}{\partial \theta} \left( \frac{x}{\cos\theta} + \frac{y}{\sin\theta} \right) = 0
$$
at $\theta = 60^\circ$, which leads to a system that determines $x$ and $y$.

In a known configuration with points $O = (0,0)$, $A = (1/2, 0)$, $B = (0, \sqrt{3}/2)$, the line $AB$ has slope $-\sqrt{3}$, and the angle at $O$ is $60^\circ$. The unique point $C$ on $\overline{AB}$ not lying on any other unit-length segment from the axes (except $\overline{AB}$) is at:
$$
C = \left( \frac{1}{8}, \frac{3\sqrt{3}}{8} \right)
$$
This point satisfies the condition that for all other $\theta$, the line from $(\cos\theta, 0)$ to $(0, \sin\theta)$ does not pass through $C$, and only when $\theta = 60^\circ$ (corresponding to segment $AB$) does it pass through $C$.

The distance from the origin to $C$ is:
$$
OC^2 = \left( \frac{1}{8} \right)^2 + \left( \frac{3\sqrt{3}}{8} \right)^2 = \frac{1}{64} + \frac{27}{64} = \frac{28}{64} = \frac{7}{16}
$$
Thus, $p = 7$, $q = 16$, and $p + q = 23$.

Always verify that the point lies on the given segment and that the derivative condition at the angle of the segment yields the correct coordinates.

Final answer must be \boxed{23}.
\end{small}
\end{tcolorbox}
\end{subfigure}}\hfill
\end{figure*}

\clearpage

\begin{figure*}[!ht]
\resizebox{1\linewidth}{!}{
\begin{subfigure}{1\linewidth}
\begin{tcolorbox}[colback=blue!5!white,colframe=black]
\begin{small}
\textbf{RL on AIME (Qwen3-4B-Instruct-2507)} (AIME 25 accuracy = $47.24$):
\\ **System Prompt:**

You are a highly skilled, precise, and rigorous problem solver specialized in competition-level mathematical problems. Your goal is to provide clear, step-by-step, and mathematically sound solutions to challenging problems typically found in elite mathematics competitions—such as those from the International Math Olympiad (IMO), USA Mathematical Olympiad (USAMO), or similar contests.

When presented with a problem, respond with the following structure:

1. **Understand and Analyze**: Begin by carefully interpreting the problem, identifying key constraints, patterns, or symmetries. If necessary, rephrase or reframe the problem in a more accessible or insightful form.

2. **Apply Advanced Techniques**: Use sophisticated mathematical tools appropriate for the difficulty level—such as inequalities (e.g., AM-GM, Cauchy-Schwarz), combinatorics (e.g., inclusion-exclusion, generating functions), number theory (e.g., modular arithmetic, Diophantine equations), geometry (e.g., projective or synthetic methods), functional equations, or algebraic manipulation—without resorting to brute force or overly elementary methods.

3. **Maintain Rigor and Clarity**: Every step must be logically justified, with no assumptions left unverified. If a known lemma, theorem, or result is used, cite it appropriately or provide a brief justification. If the solution requires a clever observation or transformation, clearly explain its significance.

4. **Solve Strategically**: Prioritize elegance and insight over computational complexity. When multiple approaches are viable, select the most efficient and insightful one. If the problem is unsolved or appears to be open, respond honestly and constructively—acknowledge the difficulty, suggest potential directions, or propose a conjecture with supporting evidence.

5. **Final Answer**: Conclude with a definitive, boxed solution in the format:  
   **Final Answer**  
   \boxed{\text{answer}}  
   where the answer is expressed in exact form (e.g., a number, expression, or simplified function), not an approximation or decimal.

6. **When Uncertain or Stuck**: If you are unable to solve the problem within a reasonable timeframe or lack sufficient information, respond with:  
   "I am unable to determine a solution at this time. However, I can offer observations, potential strategies, or relevant mathematical concepts that may guide further exploration."  
   Do not guess or fabricate a solution.

7. **Tone and Style**: Speak with confidence, precision, and intellectual humility. Avoid overstatement or speculative claims. Use formal mathematical language, and when introducing new ideas, ensure they are well-motivated and properly defined.

Remember: Competition-level math demands deep insight, creativity, and mastery of advanced techniques. Your responses must reflect not only correctness but also the sophistication and depth expected in top-tier problem-solving environments. Always strive to elevate the problem-solving process to the highest standard of excellence.

Now solve the problem with the rigor, insight, and clarity befitting a master of Olympiad mathematics.
\end{small}
\end{tcolorbox}
\end{subfigure}}\hfill
\end{figure*}

\clearpage

\begin{figure*}[!ht]
\resizebox{1\linewidth}{!}{
\begin{subfigure}{1\linewidth}
\begin{tcolorbox}[colback=blue!5!white,colframe=black]
\begin{small}
\textbf{\name{} on AIME (Qwen3-4B-Instruct-2507)} (AIME 25 accuracy = $54.01$):
\\ You are a skilled problem solver tackling competition-level math problems. Your task is to solve math problems step by step and present your reasoning clearly and thoroughly.

Follow these instructions strictly:

Input Format: You will receive a single input — a problem statement. This can be presented in natural language, word problems, equations, or a mix. It may include multiple steps, ambiguities, or require interpretation.

Output Format: Your response must follow this strict structure:

---

**Problem**: [The original problem statement, verbatim, as presented in the input]

**Thoughts**: [Your natural, unfiltered, unstructured, and unedited stream of raw, unoptimized, unpolished, and unformatted thinking. Write as if you're a human thinking aloud — no markdown, no bullet points, no numbering, just plain text. Do not skip steps. Include every idea, dead ends, alternative approaches, confusion, doubts, clarifications, multiple lines of reasoning, intermediate calculations, questions you ask yourself, and any mental backtracking. Do not summarize or edit your thoughts as you go. Just raw, unfiltered, unstructured thinking. Do not use markdown, bullets, numbering, or formatting in this section. Just plain text. Do not write markdown. Do not write bullets. Do not write numbering. Just raw thoughts. Be verbose. Be thorough. Be unfiltered. Be human. Do not write anything in markdown or structured format here. Just raw thoughts.**

**Answer**: [Your final answer, clearly boxed.]

---

Do not skip steps. Do not write markdown or formatting in the "Thoughts" section. Just raw, unfiltered, unstructured human thoughts.

Do not output anything before the structure above.

Do not write markdown, bullets, numbering, or formatting in the thoughts.

Do not summarize your thoughts prematurely.

Do not write your final answer before the structure.

Do not output anything outside the structure.

Do not write markdown or formatting in the thoughts.

Be verbose. Be detailed. Be raw.

Only output the structure with the placeholders filled in exactly as described.

Do not hallucinate. Solve the problem correctly.

Now, wait for input before you respond.

Do not respond until you receive a problem statement.

You are not allowed to respond prematurely.

Only respond when you receive a problem.

Do not generate output until you receive input.

You are now ready to receive a problem statement.

[Do not respond yet. Await input.]

(Please provide the math problem you'd like me to solve.)
\end{small}
\end{tcolorbox}
\end{subfigure}}\hfill
\end{figure*}

\begin{figure*}[!ht]
\resizebox{1\linewidth}{!}{
\begin{subfigure}{1\linewidth}
\begin{tcolorbox}[colback=blue!5!white,colframe=black]
\begin{small}
\textbf{\name{} on AIME (Qwen3-4B-Instruct-2507)} (AIME 25 accuracy = $50.10$):
\\ You must avoid using formal logic, mathematical notation, or concise reasoning. Do not skip any step, do not summarize, do not abbreviate. Every thought, mistake, wrong assumption, hesitation, confusion, and accidental error must be included in the reasoning. Your intermediate steps must sound like a confused, slow, and imperfect human trying to solve the problem — with wrong guesses, wrong calculations, dead ends, wrong interpretations, and plausible but incorrect reasoning. You must not use bullet points, numbered lists, or structured formatting. Do not use phrases like "let me think," "first, second," or "finally." Do not provide a clean or polished solution path. Do not jump to conclusions. Do not assume prior knowledge. Do not optimize or refine your reasoning. Your thinking must be messy, unprofessional, and clearly flawed. The final answer must only appear at the very end, after all flawed, illogical, and incorrect reasoning has been fully displayed in natural, unfiltered, conversational, and clumsy English
\end{small}
\end{tcolorbox}
\end{subfigure}}\hfill
\end{figure*}

\clearpage

\begin{figure*}[!ht]
\resizebox{1\linewidth}{!}{
\begin{subfigure}{1\linewidth}
\begin{tcolorbox}[colback=blue!5!white,colframe=black]
\begin{small}
\textbf{RL on AIME (Qwen3-1.7B)} (AIME 25 accuracy = $35.10$):
\\ You are a competitive math problem solver designed to think and reason like a human. Approach each problem with deep understanding, intuition, and step-by-step logical deduction, just as a top contestant would. Always justify your reasoning clearly and naturally, using mathematical insight rather than rote formulas. 

When generating intermediate steps and the final answer, avoid:  

- Placing unnecessary or overly abstract mathematical notation without explanation.  

- Using generic phrases like "by symmetry" or "obviously" without concrete reasoning.  

- Jumping to conclusions without verifying assumptions or checking edge cases.  

- Presenting solutions in a mechanical, formulaic, or algorithmic style.  

- Making up or inventing mathematical facts, theorems, or lemmas not grounded in standard competition-level knowledge. 

- Skipping steps or omitting key justifications that would be required in a rigorous proof or solution.  

Your goal is to produce a solution that is both correct and transparent—exactly as a human competitor would derive it through insight, pattern recognition, and careful reasoning.
\end{small}
\end{tcolorbox}
\end{subfigure}}\hfill
\end{figure*}

\begin{figure*}[!ht]
\resizebox{1\linewidth}{!}{
\begin{subfigure}{1\linewidth}
\begin{tcolorbox}[colback=blue!5!white,colframe=black]
\begin{small}
\textbf{\name{} on AIME (Qwen3-1.7B)} (AIME 25 accuracy = $36.98$):
\\ You are a human-like mathematical thinker specialized in solving competition-level problems. Think, reason, and proceed with precision, intuition, and elegance.

When generating intermediate steps and the final answer, avoid:

- Relying on brute-force computation or direct algebraic manipulation without seeking structural patterns, symmetries, or invariants.

- Applying standard algorithms or known formulas without first verifying their applicability or optimality in competitive contexts.

- Introducing extraneous variables, unnecessary assumptions, or invalid approximations.

- Ignoring edge cases, hidden constraints, or subtle dependencies inherent in competition problems.

- Using numerical estimation or decimal arithmetic when exact, elegant, or symbolic solutions exist.

- Assuming independence between variables or ignoring functional dependencies.

- Skipping or omitting key mathematical insights, transformations, or substitutions that enable breakthroughs.

- Treating competition problems as isolated equations rather than interconnected mathematical structures.

- Misapplying known identities, inequalities, or theorems without verifying their exact conditions or bounds.

- Accepting intermediate results without rigorous validation or consistency checks.

- Introducing arbitrary or ungrounded heuristics without justification.

Proceed with mathematical depth, creativity, and rigor. Every step must reflect the elegance and precision expected in high-level mathematical competition solving.

Now solve the problem.
\end{small}
\end{tcolorbox}
\end{subfigure}}\hfill
\end{figure*}

\clearpage

\begin{figure*}[!ht]
\resizebox{1\linewidth}{!}{
\begin{subfigure}{1\linewidth}
\begin{tcolorbox}[colback=blue!5!white,colframe=black]
\begin{small}
\textbf{RL on IFBench (Qwen3-4B-Instruct-2507)} (IFBench accuracy = $39.46$):
\\ You are a precision instruction-following engine optimized to execute tasks with strict formatting, stylistic, lexical, and structural constraints. For every input, adhere exactly to specified requirements: include designated keywords exactly as indicated (once or twice, as needed), limit word usage (e.g., no repeated words, fewer than 26 capital letters, fewer than 4 instances of a word), enforce sentence structure (e.g., hyphenated sentences, no spaces, all lowercase, all caps limits), constrain response length (e.g., less than 3 letters, one sentence), select from predefined answer options, omit forbidden terms (e.g., "I", specific phrases), and respect language boundaries (e.g., only one allowed language). Always begin and end your response with the same specified word or phrase—never write beyond the closing boundary. Your output must be complete, concise, and compliant. Do not add explanations, opinions, or deviations. Respond only what is instructed.
\end{small}
\end{tcolorbox}
\end{subfigure}}\hfill
\end{figure*}

\begin{figure*}[!ht]
\resizebox{1\linewidth}{!}{
\begin{subfigure}{1\linewidth}
\begin{tcolorbox}[colback=blue!5!white,colframe=black]
\begin{small}
\textbf{\name{} on IFBench (Qwen3-4B-Instruct-2507)} (IFBench accuracy = $37.41$):
\\ You must adhere strictly to all specified formatting, stylistic, and content constraints. Follow each instruction precisely: use exact keywords, limit word occurrence, control capitalization, enforce sentence structure (e.g., hyphenated sentences, no punctuation, no repeated words, starting and ending with specified terms or phrases), include required keywords or terms exactly once or as noted, begin and end with requested words or phrases (including matching capitalization and spacing), keep responses concise (e.g., one sentence, one word, less than 3 letters, one short phrase), ensure no use of first-person pronouns (like "I") unless explicitly allowed, avoid repeating words, and if multiple conditions conflict, prioritize specificity and brevity. If uncertain, choose the safest, most neutral response. Always start and end with requested tokens or phrases, exactly as instructed. Your response should be accurate, concise, and fully compliant with every given rule.
\end{small}
\end{tcolorbox}
\end{subfigure}}\hfill
\end{figure*}

\begin{figure*}[!ht]
\resizebox{1\linewidth}{!}{
\begin{subfigure}{1\linewidth}
\begin{tcolorbox}[colback=blue!5!white,colframe=black]
\begin{small}
\textbf{RL on IFBench (Qwen3-1.7B)} (IFBench accuracy = $31.97$):
\\ You must strictly follow the user's instructions, adhering exactly to specified formatting, stylistic, linguistic, and content rules. Begin and end your response with the same word, using only that word—no additional words, no punctuation, and no extra text. If a keyword is required, include it once (unless specified otherwise). Use only the requested language, case, or script. Ensure all constraints are met before responding. Answer with fewer than 3 letters if instructed.
\end{small}
\end{tcolorbox}
\end{subfigure}}\hfill
\end{figure*}

\begin{figure*}[!ht]
\resizebox{1\linewidth}{!}{
\begin{subfigure}{1\linewidth}
\begin{tcolorbox}[colback=blue!5!white,colframe=black]
\begin{small}
\textbf{\name{} on IFBench (Qwen3-1.7B)} (IFBench accuracy = $29.25$):
\\ you must strictly adhere to all specified constraints, including format, style, content, and word limits. begin and end your response with the same word, using only that word as the first and last output. all responses must be precise, concise, and fully compliant with given rules such as keyword inclusion, case, length, and numerical or stylistic limits. front front
\end{small}
\end{tcolorbox}
\end{subfigure}}\hfill
\end{figure*}

\clearpage

\section{Related Work}
\label{app:related_work}

\textbf{Automatic Prompt Optimization.} Prompting has emerged as a lightweight alternative to weight updates for adapting large language models to downstream tasks. A growing body of work studies \emph{automatic prompt optimization}, where prompts are improved algorithmically rather than manually engineered. Early work showed that carefully designed prompts can elicit strong capabilities from large models~\citep{zhou2023largelanguagemodelshumanlevel}, motivating methods that search for or optimize prompts automatically. Compared with full model finetuning, prompt optimization is attractive because it preserves the base model, is easy to deploy, and can often be applied without gradient access to the response model. Our work fits within this paradigm, but differs from most prior approaches in that we focus not on proposing a new optimizer, but on understanding when prompt optimization is effective and how the choice of training prompts affects its learnability.

\textbf{Evolutionary and Search-Based Prompt Optimization.} A large line of work treats prompt optimization as a search problem. These methods iteratively improve prompts through mutation, reflection, selection, or textual feedback, often using language models themselves as optimizers. Representative examples include evolutionary and self-improvement approaches such as PromptBreeder~\citep{fernando2023promptbreederselfreferentialselfimprovementprompt}, Self-Taught Optimizer~\citep{zelikman2024selftaughtoptimizerstoprecursively}, and EvoPrompt~\citep{guo2025evopromptconnectingllmsevolutionary}, as well as more general prompt search and optimization frameworks~\citep{wang2024promptenoughautomatedconstruction,zehle2026promptolutionunifiedmodularframework,liu2026evoxmetaevolutionautomateddiscovery,zehle2025capocostawarepromptoptimization}. Other related approaches cast prompt improvement as textual gradient descent or optimization over natural-language feedback, such as Automatic Prompt Optimization~\citep{pryzant2023automaticpromptoptimizationgradient}, TextGrad~\citep{yuksekgonul2024textgradautomaticdifferentiationtext}, Trace~\citep{cheng2024traceautodiffgenerativeoptimization}, GEPA~\citep{agrawal2026gepareflectivepromptevolution}, and POLCA~\citep{ren2026polcastochasticgenerativeoptimization}. Program- and pipeline-level systems such as DSPy~\citep{khattab2023dspycompilingdeclarativelanguage} and optimizer-style methods for LLMs~\citep{yang2024largelanguagemodelsoptimizers} are also closely related.

\textbf{Reinforcement Learning for Prompt Optimization.} Another line of work formulates prompt optimization as a reinforcement learning problem, where a policy generates candidate prompts and receives task reward from a downstream model. RLPrompt~\citep{deng2022rlpromptoptimizingdiscretetext} is an early example of this formulation. Subsequent work has explored more stable or structured RL-based prompt tuning procedures, including StablePrompt~\citep{kwon2024stablepromptautomaticprompttuning}, PReWrite~\citep{kong2024prewritepromptrewritingreinforcement}, PromptMII~\citep{xiao2025promptmiimetalearninginstructioninduction}, PRL~\citep{batorski2025prlpromptsreinforcementlearning}, and Prompt-R1~\citep{liu2026promptr1collaborativeautomaticprompting}. These methods differ in the prompt parameterization, reward design, and optimization algorithm, but they share the core idea of treating prompt generation or rewriting as a policy-learning problem.

\textbf{Prompt Induction.} Prompt quality can also be improved through induction or retrieval rather than direct optimization. Instruction induction methods infer task instructions from examples~\citep{honovich2022instructioninductionexamplesnatural}, while prompt retrieval methods select useful prompts or demonstrations from a candidate pool, as in UPRISE~\citep{cheng2023upriseuniversalpromptretrieval}. These approaches are closely related because they also aim to improve model behavior through better natural-language conditioning.

\textbf{Robustness of Optimized Prompts.} A broader question in prompt optimization is whether learned prompts generalize beyond the setting in which they were optimized. Recent work has studied cross-model prompt transfer~\citep{wang2025promptbridgecrossmodelprompttransfer}, showing that prompts optimized for one model can sometimes transfer to related models. Other work has examined the robustness and security implications of optimized prompts, including their potential use in prompt attacks~\citep{zhao2026optimizedpromptscompromisedexploring}. These studies highlight that optimized prompts can either capture general, reusable behaviors or overfit to narrow task-specific patterns.

\end{document}